\documentclass[11pt,a4paper]{article}

\usepackage{liatuo}          
\usepackage{placeins}
\title{ME-Dex 1.0: Bringing Heterogeneous Tactile Sensing into World Action Modeling}
\author{\textbf{Foundation Model, Li Auto Inc.}}
\shorttitle{MachEmbodied-Dex-1.0 Technical Report}
\reportdate{September 2026}

\begin{document}

\makeLiTitle

\begin{liabstract}
World Action Models bring the predictive capabilities of video models into robot action generation, providing a rich foundation for modeling future visual states.
Tactile sensing complements this foundation with direct measurements of physical interaction. 
Some existing methods use tactile features as conditioning inputs without jointly predicting future tactile states, visual observations, and actions.
Our key insight is that tactile signals, like video, provide observations of the evolving world state and should be modeled as future observations alongside video. 
We present ME-Dex-1.0 (MachEmbodied-Dex-1.0), a unified World Action Tactile Model for joint visual, tactile, and action learning. ME-Dex-1.0 adopts a Mixture-of-Transformers architecture comprising a Video Expert, a Tactile Expert, and an Action Expert, all trained with flow matching. We use shared attention connects the experts in intermediate layers, allowing action generation to draw on learned representations of visual and tactile dynamics during joint denoising.
To support multi-source heterogeneous tactile inputs, a Canonical Hand Model and a Unified Tactile Autoencoder map tactile observations from different embodiments and sensing layouts into shared spatial and latent spaces.
To address the limited availability of paired visual, tactile, and action data, we develop the Agentic Tactile Data Engine, an agent-based data production platform. It supplements RoboTwin and DexJoCo with tactile data recorded directly from force sensors during trajectory replay in simulation. 
Experiments on the RoboTwin, DexJoCo, and ManiFeel simulation platforms, together with real robot evaluations, demonstrate improved manipulation performance using both grippers and dexterous hands equipped with tactile sensing.


\vspace{1em}
\noindent\textit{\textbf{Project Page:}\enspace\url{https://machembodied.com/ME-Dex/ME-Dex1.0.html}}

\noindent\textit{\textbf{Code Repository:}\enspace\url{https://github.com/MachEmbodied/ME-Dex-1.0}}
\end{liabstract}

\vspace{4cm}
\tableofcontents


\section{Introduction}
\label{sec:intro}

\begin{figure}[bt]
  \centering
  \includegraphics[width=\textwidth]{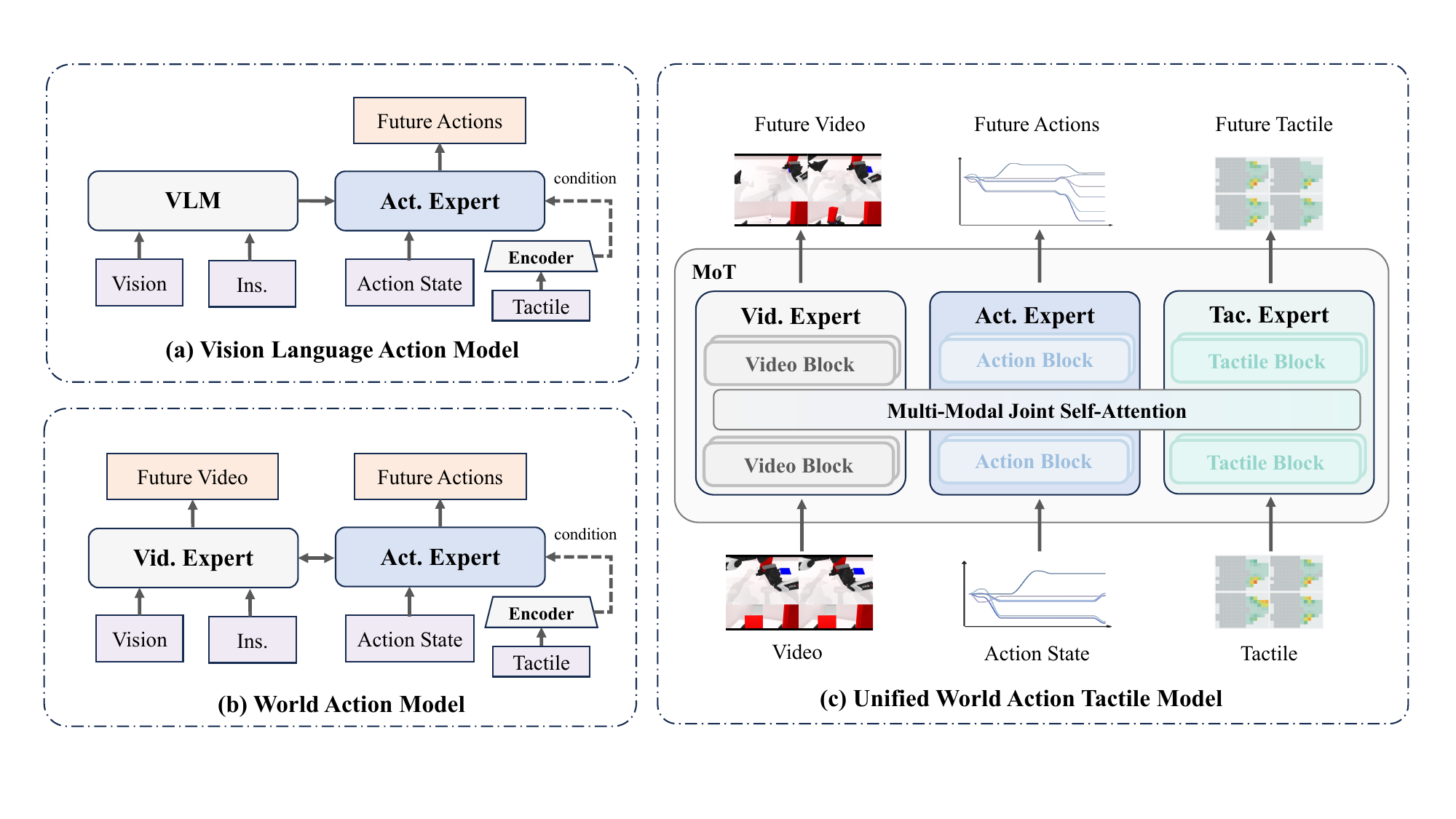}
\caption{\textbf{From tactile conditioning to unified world action tactile modeling.}
(a) Vision-language-action models generate actions from visual observations and task instructions, with some designs additionally conditioning on current tactile observations.
(b) World action models jointly predict future video and actions, optionally conditioned on current tactile observations.
(c) ME-Dex-1.0 jointly predicts future video, tactile states, and actions within a unified model.
The panels illustrate selected modeling designs rather than an exhaustive classification of existing methods.}
  \label{fig:teaser}
\end{figure}

Embodied AI aims to build agents that perceive, reason, and act through
interaction with the physical world. Recent progress in robot manipulation
has followed two closely related directions. Vision-Language-Action models
(VLAs) connect visual observations and language instructions to executable
robot actions, drawing on knowledge from pretrained vision-language models
\citep{brohan2023rt2,kim2024openvla,black2024pi0,Kim2025Openvlaoft,wang2026vlaadapter,Wang20226wvlam}. World Action Models
(WAMs) jointly learn robot actions and future visual states, bringing the
dynamics learned by video generation models into action prediction
\citep{bi2025motus, Zhun2025UWM, ye2026dreamzero,li2026lingbotva, Ye2026Gigaworld,Ma2026DECOWAM}. As illustrated in 
Figure~\ref{fig:teaser}(a--b), these approaches connect perception to control
through direct action generation or joint video-action prediction. Both seek
to turn learned representations into effective physical interaction.

Vision provides a rich foundation for this interaction by describing the
scene and how objects move. Physical interaction also depends on how the
robot applies force to an object and how the object responds. Tactile
measurements complement visual observations by describing this interaction.
For example, changes in tactile signals during a grasp can help the robot
assess whether the object is held securely. Modeling visual and tactile
dynamics together therefore
offers a path toward a more complete account of the world in which robots act.

Recent works \citep{zhao2025polytouch,heng2025vitacformer,zhu2025touchwild,huang2025tactilevla} have begun to use tactile data in robot policies
and world models. 
A common entry point is to encode tactile data as conditioning inputs, as illustrated in
Figure~\ref{fig:teaser}(a--b). This offers a straightforward way to add tactile information 
to existing models and does not require the model to predict how tactile states evolve
together with visual states and actions. Recent methods have taken steps
toward predictive tactile modeling
\citep{zheng2026omnivta,tian2026vtwam}. 
Extending this capability across robots requires both scalable tactile data and a shared predictive
model that can learn across embodiments.


We aim to build a unified World Action Tactile Model that learns from visual, action, and tactile information across embodiments. Such unification requires a single Model that aligns heterogeneous inputs through a shared Representation and supports joint training on diverse Data at scale. However, realizing this design requires addressing three gaps. The Data Gap arises because datasets such as RoboTwin~\cite{chen2025robotwin} and DexJoCo~\cite{wang2026dexjoco} lack native tactile measurements, while tactile datasets used in HATO~\cite{lin2024learning} and ManiFeel~\cite{luu2026manifeel} remain limited in scale and use different data formats. Collecting synchronized tactile and action data also requires effort. 
The Representation Gap stems from heterogeneous tactile inputs across sensors and embodiments, making them difficult to combine for joint training without a shared representation. The Model Gap concerns how to jointly learn future visual observations, tactile states, and robot actions within a predictive architecture, making future tactile representations accessible to action generation while supporting a shared representation across embodiments.

To address these gaps, we introduce ME-Dex-1.0. 
To unify data across embodiments and datasets, we design a Canonical Hand Model inspired by the human hand. We map dexterous hands and grippers to this common template and align their tactile measurements. This template establishes a shared spatial reference for heterogeneous tactile inputs. We further propose a Unified Tactile Autoencoder (AE) to encode the aligned tactile data into a shared latent space. 

At the model level, we place tactile states on equal footing with video in modeling the future world state. As shown in Figure~\ref{fig:teaser}(c), ME-Dex-1.0 adopts a three-expert Mixture-of-Transformers (MoT) architecture. The Video Expert predicts future video as a reference for action generation, while the Tactile Expert predicts future tactile latents that represent the evolution of physical interactions. The Action Expert generates action chunks from the jointly learned representations. Such three experts exchange information through shared attention. We jointly optimize all three experts for future video prediction, future tactile prediction, and action generation. This joint objective trains the Action Expert to draw on representations of future visual observations and tactile states when generating action chunks.


As for paired data, we develop the Agentic Tactile Data Engine, an agent-based platform that records tactile data from force sensors during trajectory replay in simulation. We use this platform to supplement RoboTwin and DexJoCo with tactile data aligned with their visual observations and actions.

Experiments on RoboTwin, DexJoCo, and ManiFeel show that ME-Dex-1.0 outperforms the compared baselines in average manipulation success across grippers and dexterous hands. The results also demonstrate advantages on bimanual tasks and generalization from Clean training to Random evaluation on RoboTwin.
To further assess generalization beyond simulation, we evaluate ME-Dex-1.0 on a LeRobot SO-101 with a Paxini PX6AX sensor integrated into its gripper, as well as on Xynova Flex2 dexterous hands. Experiments on both platforms show that the model can execute manipulation tasks using visual and tactile observations.

Overall, our contributions can be summarized as follows:
\begin{itemize}
    \item We present ME-Dex-1.0, a World Action Tactile Model built on a three-expert MoT architecture. Its Video Expert, Tactile Expert, and Action Expert are jointly optimized to predict future video, future tactile latents, and action chunks, allowing action generation to draw on representations of both future visual and tactile states.

    \item We propose a Canonical Hand Model inspired by the human hand and a Unified Tactile AE. These components align diverse dexterous hands and grippers to a common template and encode their tactile data into a shared latent space.

    \item We develop an Agentic Tactile Data Engine for action replay and generation, expanding RoboTwin and DexJoCo with tactile trajectories to support joint visual, tactile, and action learning.
\end{itemize}

\newpage

\section{Related Work}
\label{sec:related}

\subsection{Generative Robot Policies and World Action Models}

Generative robot policies formulate action prediction as sequence modeling. ACT \cite{zhao2023act} predicts action chunks with a conditional Transformer, while Diffusion Policy \cite{chi2023diffusion} models multimodal action distributions through iterative denoising. Large-scale datasets such as Open X-Embodiment \citep{oneill2023openx} support policy learning across tasks and embodiments. The $\pi_0$ \cite{black2024pi0} family combines a pretrained vision-language backbone with a flow-based action expert, and $\pi_{0.5}$ \cite{black2025pi05} extends this approach toward open-world generalization. X-VLA \citep{zheng2025xvla} explores learning across embodiments through soft prompts. These methods learn action generation from multimodal observations, with future environment states typically represented implicitly.

World Action Models make future visual prediction an explicit part of action learning. Motus \cite{bi2025motus} combines video generation, action generation, and semantic understanding within a Mixture-of-Transformers architecture. LingBot-VA \citep{li2026lingbotva} jointly models video and actions through an autoregressive architecture with causal attention and closed-loop observation updates. The $\tau_0$-WM model \citep{zhou2026tau0wm} also unifies video and action generation for robot learning. Fast-WAM \citep{yuan2026fastwam} studies how future video supervision supports action learning when video generation is omitted during inference. MotuBrain \citep{motubrain2026} further develops unified video-action modeling with modality-specific Transformer streams and shared attention in intermediate layers. These approaches bring the predictive capabilities of video models into robot control. Our work builds on this foundation by modeling future tactile states alongside future visual observations.

\subsection{Tactile Conditioned Robot Policies}

Tactile observations complement visual perception by providing direct information about physical interaction. Tactile-VLA \citep{huang2025tactilevla} integrates tactile signals into vision-language-action learning for contact-aware manipulation. FARM \citep{helmut2025farm} conditions a diffusion policy on tactile features and jointly predicts robot pose, gripper width, and gripping force. DECO \citep{Li2026DECO} develops a multimodal diffusion Transformer with separate conditioning pathways for visual, proprioceptive, and tactile inputs. Its plugin tactile adapter introduces tactile features through cross-attention and adapts a pretrained visual policy while keeping the original policy parameters frozen. These designs use observed tactile information to guide action generation. ManiFeel \citep{luu2026manifeel} further provides systematic comparisons across tactile modalities, encoders, and policy architectures.

Another line of work uses tactile feedback to refine policy outputs. VLA-Touch \citep{bi2025vlatouch} combines tactile feedback for high-level planning with a diffusion controller for action refinement. T-Rex \citep{niu2026trex} adopts a variable-rate Mixture-of-Transformers architecture to combine slower visuomotor processing with faster tactile feedback. Motus2 \citep{bi2026motus2} uses a lightweight tactile expert for action refinement and future force prediction, with an attention structure that prevents action-refinement tokens from attending to noised future-force tokens. These studies explore complementary roles for tactile conditioning, feedback, and predictive supervision. ME-Dex-1.0 focuses on joint visual, tactile, and action modeling, where the Action Expert interacts with intermediate representations of future visual and tactile states during joint denoising.

\subsection{Predictive Tactile Modeling}

Predictive tactile models learn the expected sensory consequences of physical interaction. Imagine2Touch \citep{ayad2024imagine2touch} predicts tactile responses from local visual observations and uses these predictions for object recognition. Visuo-Tactile World Model \citep{higuera2026vtwm} predicts future visual and tactile states conditioned on robot actions, supporting physically consistent rollouts and planning. These studies establish tactile observations as prediction targets that describe how robots interact with their environments.

Recent methods connect predictive tactile modeling with action generation and feedback control. OmniVTA \citep{zheng2026omnivta} combines a visuotactile world model with policy and tactile reflex components. TouchWorld \citep{zhou2026touchworld} separates subtask planning, visual and tactile goal prediction, action generation, and tactile residual refinement within a hierarchical system. ViTacWorld \citep{huang2026vitacworld} learns an action-conditioned visual and tactile model for trajectory generation and policy evaluation. VT-WAM \citep{tian2026vtwam} jointly learns future visual prediction, tactile deformation prediction, and action prediction through flow matching, using an asymmetric Mixture-of-Transformers architecture and contact-gated attention guidance.

Our work focuses on extending joint predictive modeling to multi-source heterogeneous tactile inputs. ME-Dex-1.0 combines a shared tactile representation across embodiments and sensing layouts with three interacting experts for video, tactile, and action modeling. The Unified Tactile Autoencoder provides a common latent space for current tactile observations and future tactile prediction, while the Mixture-of-Transformers architecture couples these predictions with action generation. This design connects tactile representation across platforms with joint learning of future observations and actions.

\subsection{Tactile Representation, Data, and Simulation}

Learning from heterogeneous tactile data requires representations that accommodate different sensors and sensing layouts. UniTouch \citep{yang2024unitouch} aligns tactile representations with semantic information from other modalities. Sparsh \citep{higuera2024sparsh} uses self-supervised pretraining to learn general-purpose representations from vision-based tactile data. AnyTouch \citep{feng2025anytouch} studies tactile representation learning across sensors and downstream tasks. Sparsh-X \citep{higuera2025sparshx} extends self-supervised tactile learning to image, audio, motion, and pressure signals from the DIGIT360 sensor. These methods explore shared representations across sensor types and complementary sensing modalities.

ME-Dex-1.0 addresses spatial correspondence across grippers and dexterous hands with different tactile sensing layouts. Our Canonical Hand Model maps tactile observations to corresponding hand regions, providing a shared spatial reference for heterogeneous three-dimensional force fields. A Unified Tactile Autoencoder then maps the aligned observations into a common latent space for tactile conditioning and future prediction. This representation connects local contact information with joint visual, tactile, and action learning across platforms.

Simulation provides a complementary source of tactile training data. TACTO \citep{wang2020tacto} renders vision-based tactile observations for robot learning, while Taxim \citep{si2021taxim} simulates optical tactile images from contact geometry. DiffTactile \citep{si2024difftactile} provides differentiable contact simulation for tactile manipulation. At the task level, RoboTwin 2.0 \citep{chen2025robotwin} supports large-scale bimanual data generation with domain randomization, DexJoCo \citep{wang2026dexjoco} provides dexterous manipulation environments and demonstrations, and ManiFeel \citep{luu2026manifeel} supports systematic studies of visuotactile policies.

To expand paired visual, tactile, and action data, we develop the Agentic Tactile Data Engine, an agent-based data production platform. It augments RoboTwin and DexJoCo, which lack native tactile observations, through trajectory replay in simulation. Tactile data are recorded directly from force sensors and paired with visual observations and actions. Together with the unified tactile representation, these data support joint learning across platforms with different embodiments and sensors.

\begin{figure}[t]
    \centering
    \includegraphics[width=\linewidth]{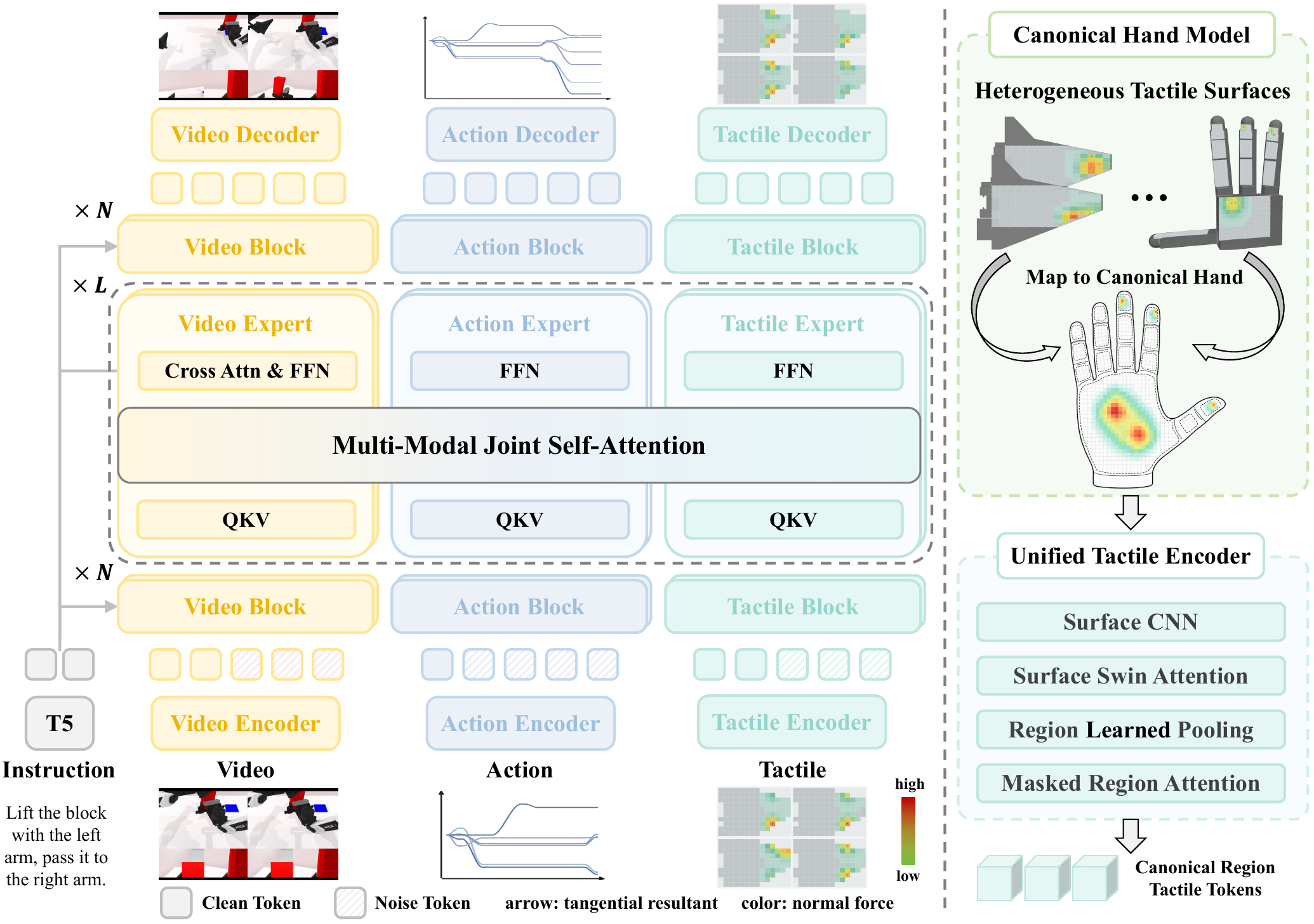}
    \caption{
\textbf{Overview of MachEmbodied-Dex-1.0}
The World Action Tactile Model (left) jointly predicts future video, tactile states, and action chunks through three experts trained with conditional flow matching.
H-Bridge shared attention connects the Video Expert, Tactile Expert, and Action Expert in intermediate layers, enabling interaction among their evolving representations during joint denoising. Shallow and deep layers retain independent processing.
The Unified Tactile Representation (right) aligns heterogeneous tactile measurements through a Canonical Hand Model and uses a shared tactile encoder to produce canonical region tokens.
}
    \label{fig:mainpipeline}
\end{figure}

\section{Method}
\label{sec:method}

\subsection{Overview}
As shown in Figure~\ref{fig:mainpipeline}, ME-Dex-1.0 jointly predicts future video, tactile states, and action chunks, treating visual and tactile signals as complementary observations of the future world state. This joint prediction allows action generation to draw on representations of both visual and tactile dynamics.

To support multi-source heterogeneous tactile data, the Unified Tactile Representation maps data from different sensors onto a Canonical Hand Model. A shared tactile autoencoder encodes the aligned data into a common latent space while preserving local force structure.

To unify visual and tactile prediction with action generation, the World Action Tactile Model adopts a Mixture-of-Transformers architecture with video, tactile, and action experts jointly trained with flow matching. H-Bridge shared attention connects the experts in intermediate layers, allowing the Action Expert to interact with representations of future visual and tactile states during joint denoising.

To address missing tactile data in existing datasets and support joint visual, tactile, and action training, we develop the Agentic Tactile Data Engine. The engine supplements existing datasets with tactile data directly from force sensors in simulation.

\subsection{Problem Formulation}

World action models jointly predict future visual observations and actions, allowing representations learned through visual prediction to inform action generation. 
ME-Dex-1.0 extends this principle to tactile prediction. Our goal is to generate action chunks informed by both future scene evolution and future physical contact.

At time $t$, the model receives a task instruction $\ell$, a visual observation $\mathbf I_t$ serving as the first video frame, a proprioceptive state $\mathbf s_t$, and the current tactile observation $\mathbf T_t$. Visual and tactile encoders map the observations into their respective latent spaces.
\begin{equation}
    \mathbf z_t^{\mathrm{vis}}=E_{\mathrm{vis}}(\mathbf I_t),
    \qquad
    \mathbf z_t^{\mathrm{tac}}=E_{\mathrm{tac}}(\mathbf T_t),
    \qquad
    \mathcal C_t=
    (\ell,\mathbf s_t,\mathbf z_t^{\mathrm{vis}},\mathbf z_t^{\mathrm{tac}}).
    \label{eq:mex_conditions}
\end{equation}
The tactile observation contains the available signals from both hands, organized according to the canonical representation defined in Section~\ref{sec:unified_tactile_representation}.

Let $\mathbf A^+=(\mathbf a_t,\ldots,\mathbf a_{t+k-1})$ denote a chunk of $k$ actions, and let $\mathbf V^+$ and $\mathbf T^+$ denote future video and tactile sequences covering the same physical time window. The superscript $+$ denotes future sequences. We represent the future observable world through two complementary modalities.
\begin{equation}
    \mathbf Z^{\mathrm{vis},+}=E_{\mathrm{vis}}(\mathbf V^+),
    \qquad
    \mathbf Z^{\mathrm{tac},+}=E_{\mathrm{tac}}(\mathbf T^+),
    \qquad
    \mathbf Z^{\mathrm{world},+}
    =\bigl(\mathbf Z^{\mathrm{vis},+},\mathbf Z^{\mathrm{tac},+}\bigr),
    \label{eq:mex_future_world}
\end{equation}
where $E_{\mathrm{tac}}$ operates frame by frame on a tactile sequence. 
Video describes visible scene evolution, while tactile observations describe the spatial distribution and temporal evolution of physical contact.
Both modalities are prediction targets in the joint model.
\begin{equation}
    p_\theta(\mathbf A^+,\mathbf Z^{\mathrm{world},+}\mid\mathcal C_t)
    =p_\theta\!\left(
        \mathbf A^+,\mathbf Z^{\mathrm{vis},+},\mathbf Z^{\mathrm{tac},+}
        \mid\mathcal C_t
    \right).
    \label{eq:mex_joint_distribution}
\end{equation}
The resulting action policy is the action marginal of this joint distribution.
\begin{equation}
    \pi_\theta(\mathbf A^+\mid\mathcal C_t)
    =\int
    p_\theta\!\left(
        \mathbf A^+,\mathbf Z^{\mathrm{vis},+},\mathbf Z^{\mathrm{tac},+}
        \mid\mathcal C_t
    \right)
    \,\mathrm d\mathbf Z^{\mathrm{vis},+}
    \,\mathrm d\mathbf Z^{\mathrm{tac},+}.
    \label{eq:mex_action_marginal}
\end{equation}
This marginal is realized by jointly sampling actions and future observation latents and retaining the action output. 
Future video and tactile states participate in the same generative process as the actions. Their ground-truth values are used as training targets and are unavailable at inference.

To make this dependence part of the computation, the three experts exchange intermediate representations during joint denoising. Let $\mathbf H_j^m$ denote the hidden features of modality $m\in\{\mathrm{vis},\mathrm{tac},\mathrm{act}\}$ at layer $j$, and let $\mathcal B$ denote the set of H-Bridge layers. The action update at a bridge layer can be expressed abstractly as
\begin{equation}
    \mathbf H_{j+1}^{\mathrm{act}}
    =\mathcal F_j^{\mathrm{act}}\!\left(
        \mathbf H_j^{\mathrm{act}},
        \mathbf H_j^{\mathrm{vis}},
        \mathbf H_j^{\mathrm{tac}}
    \right),
    \qquad j\in\mathcal B,
    \label{eq:mex_action_interaction}
\end{equation}
where $\mathcal F_j^{\mathrm{act}}$ denotes the action-side Transformer update with shared attention. The visual and tactile features evolve as their experts estimate the corresponding future observations. Action generation therefore interacts with representations of both futures throughout the shared layers and successive denoising steps. This provides a direct computational path through which learning future contact dynamics, alongside visual dynamics, can inform the action prediction.

\newpage

\subsection{Unified Tactile Representation}
\label{sec:unified_tactile_representation}

\begin{figure}[bt]
  \centering
  \includegraphics[width=\textwidth]{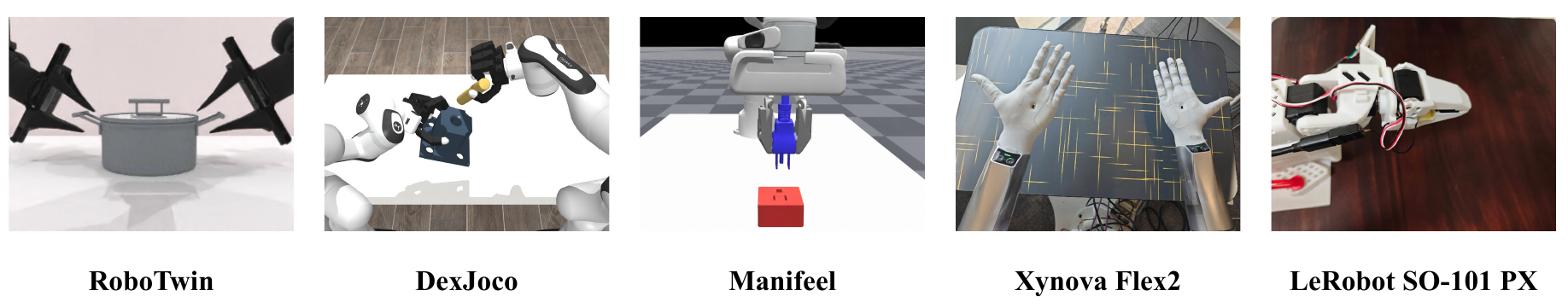}
  \caption{
\textbf{Multi-source heterogeneous tactile inputs.}
From left to right, RoboTwin, DexJoCo, ManiFeel, Xynova Flex2 dexterous hands, and a tactile-equipped LeRobot SO-101 gripper.
These sources motivate MachEmbodied-Dex-1.0's unified tactile representation, which expresses heterogeneous contact measurements across embodiments and sensing layouts in a common spatial and latent space.
}
  \label{fig:list_of_env}
\end{figure}

\subsubsection{Tactile Inputs}

Our tactile inputs come from diverse grippers and dexterous hands, including grippers in RoboTwin and ManiFeel, dexterous hands in DexJoCo, and Xynova Flex2 dexterous hands, as described in Figure~\ref{fig:list_of_env}. These examples span different sensing surfaces and layouts. To model their contact dynamics with a shared predictor, we need a representation that retains the spatial structure of local contact while providing a common interface across end effectors.

Each sensing surface $r$ provides a force field $\mathbf F_t^{(r)}\in\mathbb R^{3\times H\times W}$ containing local normal and two tangential force components on an $H\times W$ grid. A layout mask identifies valid sensing positions, including positions without contact. Preprocessing applies a smooth dead zone, channelwise $\operatorname{asinh}$ compression, normalization, and clipping. Normal and tangential components lie in $[0,1]$ and $[-1,1]$, respectively. 


\subsubsection{Canonical Hand Model}

To give tactile measurements a consistent spatial meaning across embodiments, we design a Canonical Hand Model inspired by the human hand and treat grippers and dexterous hands as morphological variations of this template. The template defines corresponding finger and palm regions for the left and right hands. Region assignments establish a common spatial reference without requiring identical surface geometry across end effectors.

Figure~\ref{fig:mainpipeline} illustrates the mapping. Each dexterous-hand sensing surface is assigned to its corresponding finger or palm region. The two sensing surfaces of each RoboTwin gripper map to thumb and index regions, with each surface divided into four disjoint, ordered patches. Different sources observe different subsets of the template. A region mask distinguishes unavailable regions from observed regions without contact, preserving the difference between missing measurements and measured absence of contact.


\subsubsection{Unified Tactile Autoencoder}

Canonical assignment establishes spatial correspondence, but the underlying force fields still differ in resolution, surface geometry, and sensing coverage. A shared future-tactile predictor therefore needs a compact latent representation with consistent spatial organization across these layouts. The representation must also preserve local force patterns and contact transitions, since these are the physical quantities that future prediction should capture. We address these requirements with a Unified Tactile Autoencoder that learns to reconstruct observed force fields from canonical region tokens, providing the tactile counterpart to the latent representation used for video prediction.

The deterministic encoder shares parameters across sensing surfaces and processes each frame independently. A residual convolutional network extracts local force features, followed by window and shifted-window attention within each surface. Masked learned pooling aggregates features over valid sensing positions. Gripper surfaces are pooled separately within their four patches, while dexterous-hand surfaces use whole-surface pooling.

The tokens are assigned to canonical slots and augmented with region and pooling-type embeddings. Two masked Transformer layers exchange information among valid regions within a frame, producing $\mathbf z_t^{\mathrm{tac}}\in\mathbb R^{N\times d}$ with $N$ region tokens of width $d$. Unobserved regions remain masked and have zero output. Shared parameters and consistent region ordering provide a common latent representation across embodiments, while region tokens retain the spatial organization needed to model local contact evolution.

The decoder gathers the tokens assigned to each surface and uses local grid positions as queries in cross attention. A residual multilayer perceptron and convolutional refinement recover surface features. A force head predicts normal and tangential components, while a contact head predicts contact probabilities. The reconstructed force field is the product of the force prediction, contact probability, and valid-position mask, with the latter two broadcast across force channels.

We train the autoencoder with
\begin{equation}
    \mathcal L_{\mathrm{AE}}
    =\mathcal L_{\mathrm{force}}
    +\mathcal L_{\mathrm{contact}}
    +\mathcal L_{\mathrm{change}}
    +\lambda_{\mathrm{lat}}\mathcal L_{\mathrm{latent}}.
    \label{eq:mex_ae_objective}
\end{equation}
The force loss applies Smooth L1 supervision to force predictions before contact gating at contact positions and to gated reconstructions at background positions. The contact loss uses balanced binary cross entropy with soft labels derived from force magnitudes before dead-zone processing. The change loss matches force differences between adjacent frames, with greater weight on contact onset and release. The latent loss constrains pooled features of valid regions without contact toward those obtained from zero input under the same layout, weighted by $\lambda_{\mathrm{lat}}$. These terms encourage the representation to preserve force magnitude, contact support, and temporal changes while giving observed non-contact regions a consistent latent reference. After pretraining, the autoencoder is frozen and supplies observed tactile conditions and future tactile targets for the joint model, as well as the decoder for predicted tactile latents.


\subsection{World Action Tactile Model}
\label{sec:world_action_tactile_model}

\subsubsection{Three-Expert Architecture}

The World Action Tactile Model uses a Mixture-of-Transformers architecture with a Video Expert, a Tactile Expert, and an Action Expert. Each expert is a separate Transformer trained with flow matching. The Video Expert follows Wan2.2-5B~\citep{wan2025wan}, with 30 Transformer layers and a hidden width of 3072. The Action Expert and Tactile Expert also have 30 layers, with hidden widths of 1024 and 512, respectively. We initialize the Video Expert and Action Expert from Motus weights pretrained on diverse robot data~\citep{bi2025motus} and initialize the Tactile Expert randomly.

During training, the Video Expert receives the encoded first frame together with noisy future video latents; the Action Expert receives the proprioceptive state together with noisy ground-truth actions; and the Tactile Expert receives the encoded current tactile frame together with noisy future tactile latents. The corresponding ground-truth future video, actions, and tactile observations provide supervision through their flow-matching targets. Each expert retains its own representation and hidden width, while their common depth supports interaction at intermediate layers.

The Tactile Expert treats tactile as a temporal observation modality with a structure analogous to video. A future tactile sequence contains $T$ frames, each describing the available tactile signals from both hands at its corresponding time. Applying the frozen tactile encoder frame by frame produces
\begin{equation}
    \mathbf Z^{\mathrm{tac},+}
    =\bigl(
        E_{\mathrm{tac}}(\mathbf T_1^+),\ldots,
        E_{\mathrm{tac}}(\mathbf T_T^+)
    \bigr)
    \in\mathbb R^{T\times N\times d}.
    \label{eq:mex_tactile_sequence}
\end{equation}
The Tactile Expert is a diffusion Transformer (DiT) that denoises this sequence in latent space, conditioned on the current tactile frame and coupled to the Video and Action Experts. At inference, the future sequence is initialized as Gaussian noise with the same latent shape. After joint denoising, the tactile decoder maps each predicted latent frame back to tactile force fields. 
This design models tactile observations as a temporal sequence of spatially organized latents. Predicting this sequence brings contact dynamics into world action modeling through the same generative formulation as video.

\vspace{-3mm}
\subsubsection{H-Bridge Shared Attention}

Inspired by H-Bridge~\citep{wang2026hbridge}, we restrict cross-modal interaction to an intermediate set of Transformer layers $\mathcal B$. Within these layers, the experts project their features into compatible attention dimensions and attend jointly over video, tactile, and action tokens before mapping the outputs back to their own hidden widths. In the shallow and deep layers, each expert processes its tokens independently. This design concentrates information exchange in the middle of the network, preserves modality-specific processing at its input and output stages, and reduces the number of layers requiring cross-modal attention. The Action Expert can thus access representations of both future visual evolution and contact dynamics through the shared layers.

\vspace{-3mm}
\subsubsection{Joint Training and Inference}

We train all three experts with conditional flow matching. For modalities $m\in\mathcal M=\{\mathrm{vis},\mathrm{tac},\mathrm{act}\}$, the clean future targets are
\begin{equation}
    \mathbf x_0^{\mathrm{vis}}=\mathbf Z^{\mathrm{vis},+},
    \qquad
    \mathbf x_0^{\mathrm{tac}}=\mathbf Z^{\mathrm{tac},+},
    \qquad
    \mathbf x_0^{\mathrm{act}}=\mathbf A^+.
    \label{eq:mex_clean_targets}
\end{equation}
Each modality independently samples a noise level $\sigma_m\in[0,1]$ and standard Gaussian noise $\boldsymbol\epsilon^m$. We construct noisy targets and their target velocities as
\begin{equation}
    \mathbf x_{\sigma_m}^m
    =(1-\sigma_m)\mathbf x_0^m+\sigma_m\boldsymbol\epsilon^m,
    \qquad
    \mathbf u^m=\boldsymbol\epsilon^m-\mathbf x_0^m.
    \label{eq:mex_flow_interpolation}
\end{equation}
The first video frame, proprioceptive state, and current tactile frame remain uncorrupted conditions. Let $\mathbf X_{\boldsymbol\sigma}=\{\mathbf x_{\sigma_m}^m\}_{m\in\mathcal M}$ and $\boldsymbol\sigma=\{\sigma_m\}_{m\in\mathcal M}$. The experts predict modality-specific velocities while exchanging features through H-Bridge, giving the loss
\begin{equation}
    \mathcal L_{\mathrm{FM}}^m
    =\mathbb E\!\left[
        \ell_m\!\left(
            \mathbf v_\theta^m(
                \mathbf X_{\boldsymbol\sigma},
                \boldsymbol\sigma,
                \mathcal C_t
            ),
            \mathbf u^m
        \right)
    \right].
    \label{eq:mex_flow_objective}
\end{equation}
The expectation is over demonstration windows, noise levels, and Gaussian noise. For video and actions, $\ell_m$ is mean squared error. For tactile prediction, it is a weighted squared error that emphasizes changes in ground-truth tactile latents. The joint objective is
\begin{equation}
    \mathcal L_{\mathrm{joint}}
    =\lambda_{\mathrm{vis}}\mathcal L_{\mathrm{FM}}^{\mathrm{vis}}
    +\lambda_{\mathrm{tac}}\mathcal L_{\mathrm{FM}}^{\mathrm{tac}}
    +\lambda_{\mathrm{act}}\mathcal L_{\mathrm{FM}}^{\mathrm{act}},
    \label{eq:mex_joint_objective}
\end{equation}
where the $\lambda$ coefficients balance the three prediction tasks. Future visual and tactile supervision shapes the representations exchanged with the Action Expert, while action supervision trains the coupled model to generate the demonstrated controls. Tactile dynamics thus participate in both future observation prediction and the intermediate computation of actions.

At inference, future video latents, future tactile latents, and actions are initialized from Gaussian noise and jointly updated from noise levels of one toward zero using the predicted velocities, under fixed observation conditions $\mathcal C_t$. H-Bridge exchanges information among the evolving modality representations during these updates. The resulting action chunk provides the control output, and the frozen visual and tactile decoders can recover the corresponding predicted observations from their latents.


\vspace{-3mm}
\subsection{Agentic Tactile Data Engine}

We develop an Agentic Tactile Data Engine for action replay and generation in simulation. Replaying existing RoboTwin and DexJoCo actions allows force sensors in the simulation to directly record tactile signals for those trajectories. 
Generating and executing new action trajectories provides additional tactile data.
The resulting trajectories pair tactile signals with visual observations and actions over shared time windows, providing both current observation conditions and aligned future targets for joint learning.

\vspace{-3mm}
\section{Experiments}


\subsection{Experimental Setup}

\textbf{Datasets and Platforms.}
\textit{RoboTwin.} We augment RoboTwin demonstrations with tactile signals collected directly from force sensors in simulation. We train jointly on 50 tasks using 27,500 Clean and Random trajectories and evaluate under both configurations. An additional setting uses only 2,500 Clean trajectories to assess generalization to Random configurations.

\textit{DexJoCo.} We augment DexJoCo demonstrations with tactile signals collected directly from force sensors in simulation. Following its official multi-task setting, we jointly train on 1,100 demonstrations spanning six single-arm and five bimanual tasks~\citep{wang2026dexjoco}.

\textit{ManiFeel.} ManiFeel provides three-dimensional tactile force observations~\citep{luu2026manifeel}. We use 50 demonstrations per task across four insertion tasks, training and evaluating a separate policy for each. 

\textit{Physical Robots.} We qualitatively assess real-world deployment on two robot systems: a single-arm LeRobot SO-101 with a Paxini PX6AX tactile sensor integrated into its gripper, and a dual-arm robot equipped with XR5-10-FS robotic arms and Xynova Flex2 dexterous hands.


\textbf{Baselines and Evaluation.}
On RoboTwin, we compare against $\pi_{0.5}$~\citep{black2025pi05}, Fast-WAM~\citep{yuan2026fastwam}, and Motus~\citep{bi2025motus}.
On DexJoCo, we compare against DP-T \cite{chi2023diffusion}, $\pi_{0.5}$, DECO and DECO.p~\citep{Li2026DECO}. The DP-T results are quoted from DexJoCo~\citep{wang2026dexjoco}, the $\pi_{0.5}$ results are obtained by training the model use the use the official configuration, the DECO results follow its official benchmark release~\citep{Li2026DECO}, and the results of DECO.p follows DECO’s official force–tactile approach with its released DexJoCo checkpoint on our tactile-augmented DexJoCo raw data. 
On ManiFeel, we use the official policy as our baseline and denote it as ManiFeel.
To examine the effect of tactile observations on different policies, we construct tactile-conditioned variants of $\pi_{0.5}$ and Fast-WAM, denoted as $\pi_{0.5}$+\textsc{Tac} and Fast-WAM+\textsc{Tac}, where
we encode tactile input using the pretrained Unified Tactile Autoencoder, for $\pi_{0.5}$+\textsc{Tac}, we concatenate the resulting tactile tokens with the image and language tokens as prefix tokens that condition the action expert, for Fast-WAM+\textsc{Tac} and ME-Dex-1.0(Tac Cond.) we concat them with the language tokens generated from T5 \citep{raffel2020exploring}.
The remaining architecture and training settings are unchanged from the corresponding baseline. For the policies trained in this study, we use a single training seed for each experiment. Our primary manipulation metric is closed-loop task success rate, with success determined by the official criteria of each benchmark.



\textbf{Implementation Details.}
We initialize the Video Expert and Action Expert of ME-Dex-1.0 from pretrained Motus weights~\citep{bi2025motus} and randomly initialize the Tactile Expert.
The Unified Tactile Autoencoder is pretrained separately on tactile data from multiple sources and remains frozen during policy training on each benchmark to prevent the tactile representation from drifting across tasks.

\subsection{Quantitative Comparison}
\subsubsection{RoboTwin Benchmark}

\textbf{Mixed Clean and Random Training.}
Under mixed Clean and Random training, ME-Dex-1.0 achieves success rates of 91.56\% and 91.92\% in the Clean and Random configurations, respectively, as shown in Table~\ref{tab:robotwin_mixed_results}.
These results exceed Fast-WAM by 2.34 and 2.70 percentage points and Fast-WAM+\textsc{Tac} by 1.34 and 1.70 percentage points.

\begin{table*}[t]
\centering
\caption{
\textbf{Multi-task manipulation performance on RoboTwin.}
Success rates (\%) under Clean + Random training.
Train and Eval indicate current tactile observation usage.
Zeroed denotes zero tactile input at evaluation.
Future tactile prediction denotes training with future tactile targets and predicting future tactile states during inference.
VA models video and actions with joint attention at all layers.
Tac Cond. adds current tactile conditions, and Full Joint additionally predicts future tactile states.
The complete model uses H-Bridge shared attention.
}
\label{tab:robotwin_mixed_results}
\small
\setlength{\tabcolsep}{3pt}
\renewcommand{\arraystretch}{1.08}
\begin{tabularx}{\linewidth}{@{}l *{6}{>{\centering\arraybackslash}X}@{}}
\toprule
\multirow[c]{2}{*}[-0.5ex]{Method}
& \multicolumn{2}{c}{Tactile obs.}
& \multirow[c]{2}{*}[-0.5ex]{%
    \shortstack{Future tactile\\prediction}}
& \multirow[c]{2}{*}[-0.5ex]{H-Bridge}
& \multicolumn{2}{c}{Success rate $\uparrow$} \\
\cmidrule(lr){2-3}
\cmidrule(l){6-7}
& Train & Eval & & & Clean & Random \\
\midrule
\multicolumn{7}{@{}l}{\textit{Baselines}} \\
$\pi_{0.5}$
& No & Not used & No & --- & 52.76 & 60.74 \\
Fast-WAM
& No & Not used & No & --- & 89.22 & 89.22 \\
Motus
& No & Not used & No & --- & 88.66 & 87.02 \\
$\pi_{0.5}$+\textsc{Tac}
& Yes & Observed & No & --- & 52.60 & 57.66 \\
Fast-WAM+\textsc{Tac}
& Yes & Observed & No & --- & 90.22 & 90.22 \\
\midrule
\multicolumn{7}{@{}l}{\textit{ME-Dex-1.0 and variants}} \\
ME-Dex-1.0 (VA)
& No & Not used & No & No & 88.74 & 86.90 \\
ME-Dex-1.0 (Tac Cond.)
& Yes & Observed & No & No & 89.04 & 89.54 \\
ME-Dex-1.0 (Full Joint)
& Yes & Observed & Yes & No & 89.36 & 90.60 \\
\textbf{ME-Dex-1.0}
& Yes & Observed & Yes & Yes & 91.56 & \textbf{91.92} \\
ME-Dex-1.0 (Zero Tac Input)
& Yes & Zeroed & Yes & Yes & \textbf{91.70} & 91.74 \\
\bottomrule
\end{tabularx}
\end{table*}

\begin{table}[t]
\centering
\caption{
\textbf{Manipulation performance on RoboTwin under Clean training.}
Baseline results are taken from the RoboTwin leaderboard~\citep{chen2025robotwin}.
Average denotes the mean success rate across the two evaluation configurations.
Tactile observations are set to zero for ME-Dex-1.0 under both evaluation configurations.}
\label{tab:robotwin_clean2random}
\small
\setlength{\tabcolsep}{4pt}
\renewcommand{\arraystretch}{1.15}
\begin{tabularx}{\linewidth}{@{}l *{3}{>{\centering\arraybackslash}X}@{}}
\toprule
\multirow[c]{2}{*}[-0.5ex]{Method}
& \multicolumn{3}{c}{Success rate (\%) $\uparrow$} \\
\cmidrule(l){2-4}
& \mbox{Clean$\rightarrow$Clean}
& \mbox{Clean$\rightarrow$Random}
& Average \\
\midrule
Fast-WAM & 77.8 & 1.9 & 39.9 \\
Xiaomi Robotics-0 & 62.9 & 18.2 & 40.5 \\
$\pi_{0.5}$ & 70.7 & 46.0 & 58.4 \\
4D-WAM & 81.5 & 41.8 & 61.6 \\
GigaBrain-0.7 & 66.8 & 67.9 & 67.3 \\
OLA-Sem & 75.1 & 67.6 & 71.3 \\
\midrule
\textbf{ME-Dex-1.0} & \textbf{89.6} & \textbf{68.1} & \textbf{78.9} \\
\bottomrule
\end{tabularx}
\end{table}

Within the ME-Dex-1.0 variants, adding current tactile conditions to the video-action model increases Random success from 86.90\% to 89.54\%.
Jointly predicting future tactile states further increases success to 90.60\%, and using H-Bridge shared attention raises it to 91.92\%.
The effect of tactile conditioning varies across external baselines. Under Random evaluation, Fast-WAM gains 1.00 percentage point, whereas $\pi_{0.5}$ loses 3.08 percentage points.

Setting the current tactile input of the trained ME-Dex-1.0 model to zero yields 91.70\% on Clean and 91.74\% on Random, close to the results obtained with observed tactile input.
This comparison shows limited sensitivity to current tactile input in this setting. The zero-input variant retains tactile supervision during training and future tactile prediction during inference.

\textbf{Training on Clean Data Only.}
We further evaluate generalization from Clean to Random configurations by training ME-Dex-1.0 on only 2,500 Clean demonstrations across the 50 tasks and setting tactile observations to zero during both Clean and Random evaluation.
As shown in Table~\ref{tab:robotwin_clean2random}, ME-Dex-1.0 achieves success rates of 89.6\% on Clean and 68.1\% on Random, with an average of 78.9\%, exceeding the strongest listed baseline, OLA-Sem, by 7.6 percentage points on average.
Its Random success rate is close to the 67.9\% reported for GigaBrain-0.7, while its Clean success rate exceeds the strongest listed Clean baseline, 4D-WAM, by 8.1 percentage points.

\subsubsection{DexJoCo Benchmark}

\begin{table*}[t]
\centering
\caption{
\textbf{Multi-task manipulation performance on DexJoCo.} Success rates (\%) under the \textit{rand-obj} configuration. DP-T, $\pi_{0.5}$, and DECO use no tactile input, whereas $\pi_{0.5}$+\textsc{Tac}, DECO.p, and ME-Dex-1.0 use tactile data. DECO.p is the tactile-adapter variant of DECO. /B denotes a bimanual task, and Bimanual avg. is the equal-weight mean across the five /B tasks. DP-T results are quoted from DexJoCo, and DECO results are taken from its official benchmark release. Bold indicates the highest success rate in each row.
}
\label{tab:multitask_performance}
\small
\renewcommand{\arraystretch}{1.15}
\setlength{\tabcolsep}{5pt}
\begin{tabularx}{\linewidth}{@{}l *{6}{>{\centering\arraybackslash}X}@{}}
\toprule
Task & DP-T & $\pi_{0.5}$ & $\pi_{0.5}$+\textsc{Tac} & DECO & DECO.p & ME-Dex-1.0 \\
\midrule
Hammer Nail  & 58.7 & 40.0 & 10.0 & 68.0 & 76.0 & \textbf{86.0} \\
Click Mouse  & 38.7 & 76.0 & 80.0 & \textbf{90.0} & 80.0 & 76.0 \\
Pick Bucket  & 55.3 & 82.0 & 46.0 & 80.0 & 84.0 & \textbf{92.0} \\
Pinch Tongs  & 6.0  & 16.0 & 0.0 & 46.0 & 60.0 & \textbf{92.0} \\
Fold Glasses & 11.3 & 24.0 & 40.0 & 50.0 & \textbf{62.0} & 56.0 \\
Water Plant  & 60.0 & 66.0 & 82.0 & \textbf{88.0} & 76.0 & 72.0 \\
\midrule
Unlock iPad/B & 0.0 & 8.0 & 0.0 & 40.0 & 32.0 & \textbf{48.0} \\
Hanoi/B      & 8.0 & 14.0 & 2.0 & 18.0 & 24.0 & \textbf{32.0} \\
Assembly/B   & 1.3 & 4.0 & 6.0 & 0.0 & 6.0 & \textbf{28.0} \\
Microwave/B  & 42.7 & 70.0 & 44.0 & 64.0 & 68.0 & \textbf{88.0} \\
Photograph/B & 28.0 & 4.0 & 22.0 & \textbf{76.0} & 68.0 & 48.0 \\
\midrule
Bimanual avg. & 16.0 & 20.0 & 14.8 & 39.6 & 39.6 & \textbf{48.8} \\
Avg. & 33.2 & 36.7 & 30.2 & 56.4 & 57.8 & \textbf{65.3} \\
\bottomrule
\end{tabularx}
\end{table*}

Under multi-task training across all 11 tasks in the \textit{rand-obj} configuration, ME-Dex-1.0 achieves the highest overall success rate of 65.3\%, as shown in Table~\ref{tab:multitask_performance}. It outperforms both baselines without tactile input and tactile-conditioned methods, including $\pi_{0.5}$+\textsc{Tac} and DECO.p, and ranks first on seven tasks.

The five bimanual tasks provide tactile data from both hands, offering a useful setting for evaluating joint visual, tactile, and action modeling during coordinated manipulation. ME-Dex-1.0 achieves a bimanual average success rate of 48.8\%, compared with 39.6\% for both DECO and DECO.p. It obtains the highest success rate on four of the five bimanual tasks, with strong performance on Assembly and Microwave. These results demonstrate its effectiveness across several tasks requiring coordinated contact between both hands, although DECO and DECO.p remain stronger on Photograph.

Among the single-arm tasks, ME-Dex-1.0 achieves the highest success rate on Pick Bucket and Pinch Tongs, reaching 92.0\% on both. Its advantages are not uniform across tasks, with DECO.p performing better on Fold Glasses and DECO leading on Click Mouse and Water Plant.




\subsubsection{ManiFeel Benchmark}

\begin{table}[t]
\centering
\caption{
\textbf{Manipulation performance on ManiFeel.}
Success rates (\%) with single-task training and 50 evaluation episodes per task.
}
\label{tab:manifeel_results}
\small
\setlength{\tabcolsep}{4pt}
\renewcommand{\arraystretch}{1.15}
\begin{tabularx}{\linewidth}{@{}l *{5}{>{\centering\arraybackslash}X}@{}}
\toprule
Method
& \shortstack{Peg\\Insertion}
& \shortstack{USB\\Insertion}
& \shortstack{Power Plug\\Insertion}
& \shortstack{Gear\\Assembly}
& Average $\uparrow$ \\
\midrule
ManiFeel & 40.0 & 56.0 & 58.0 & \textbf{66.0} & 55.0 \\
ME-Dex-1.0 & \textbf{58.0} & \textbf{74.0} & \textbf{88.0} & 60.0 & \textbf{70.0} \\
\bottomrule
\end{tabularx}
\end{table}

The ManiFeel baseline uses Diffusion Policy to generate action sequences conditioned on visual observations and three-dimensional tactile force fields.
ME-Dex-1.0 jointly predicts future video, tactile states, and action chunks, allowing the experts to exchange intermediate representations during prediction.

Table~\ref{tab:manifeel_results} shows that ME-Dex-1.0 outperforms the ManiFeel baseline on three of the four tasks. Its average success rate reaches 70.0\%, an improvement of 15.0 percentage points over the baseline.
The largest gain occurs on Power Plug Insertion, where success increases from 58.0\% to 88.0\%. Peg Insertion and USB Insertion each improve by 18.0 percentage points.
On Gear Assembly, however, success decreases from 66.0\% to 60.0\%, indicating that the gains are not uniform across tasks.

\begin{figure}[t]
    \centering
    \includegraphics[width=\linewidth]{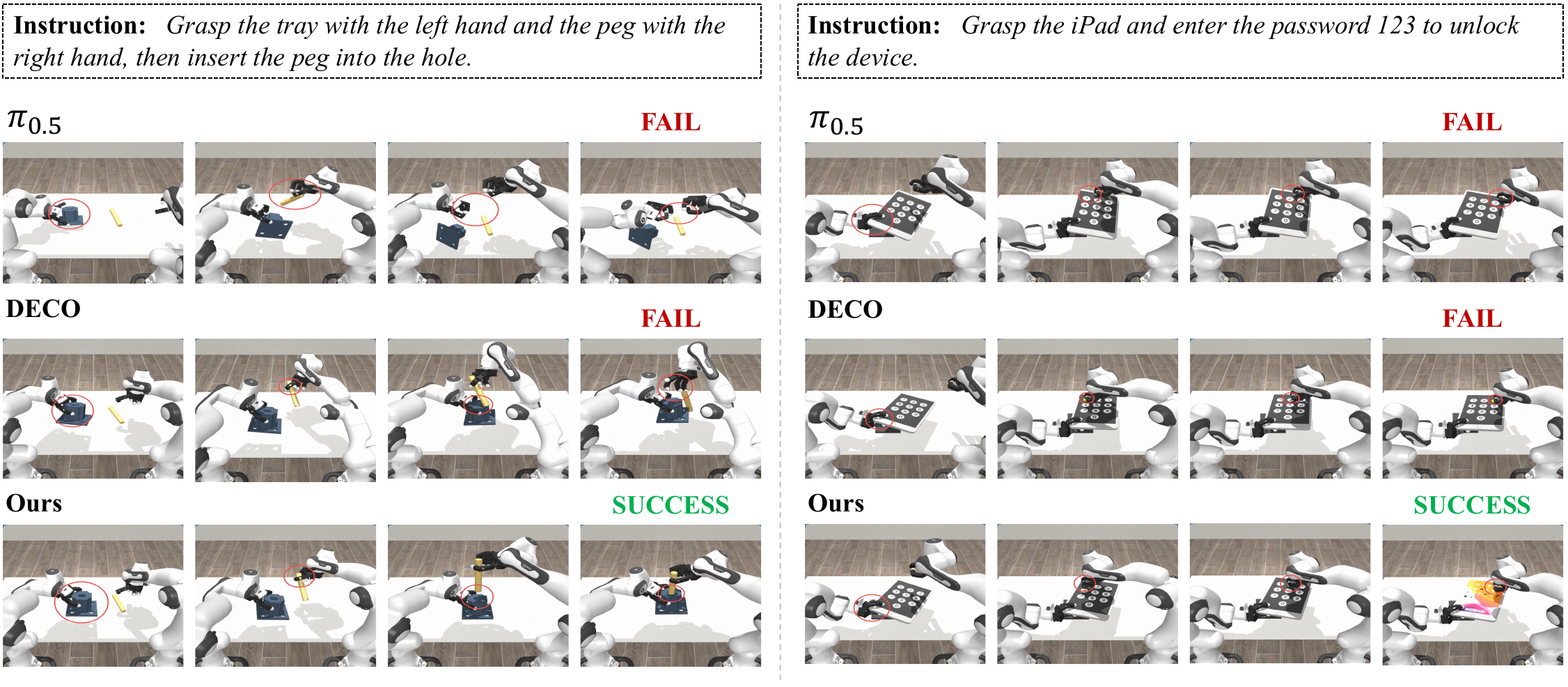}
    \caption{
\textbf{Qualitative comparisons on DexJoCo.}
Rollouts of $\pi_{0.5}$, DECO, and ME-Dex-1.0 on Assembly (left) and Unlock iPad (right). Frames progress from left to right.
}
    \label{fig:qualitative-dexjoco}
    \vspace{-3mm}
\end{figure}

\vspace{-3mm}
\subsection{Qualitative Results}
\subsubsection{DexJoco}
\vspace{-1mm}
Assembly and Unlock iPad examples highlight the benefits of ME-Dex-1.0 on contact-rich bimanual tasks. Both tasks require one hand to stabilize an object while the other performs precise interactions, providing tactile information from both hands. In the figure~\ref{fig:qualitative-dexjoco} of shown rollouts, ME-Dex-1.0 exhibits more effective coordination than $\pi_{0.5}$ and DECO during peg insertion and password entry.

\vspace{-3mm}
\subsubsection{Robotwin}
\begin{figure}[t]
    \centering
    \includegraphics[width=\linewidth]{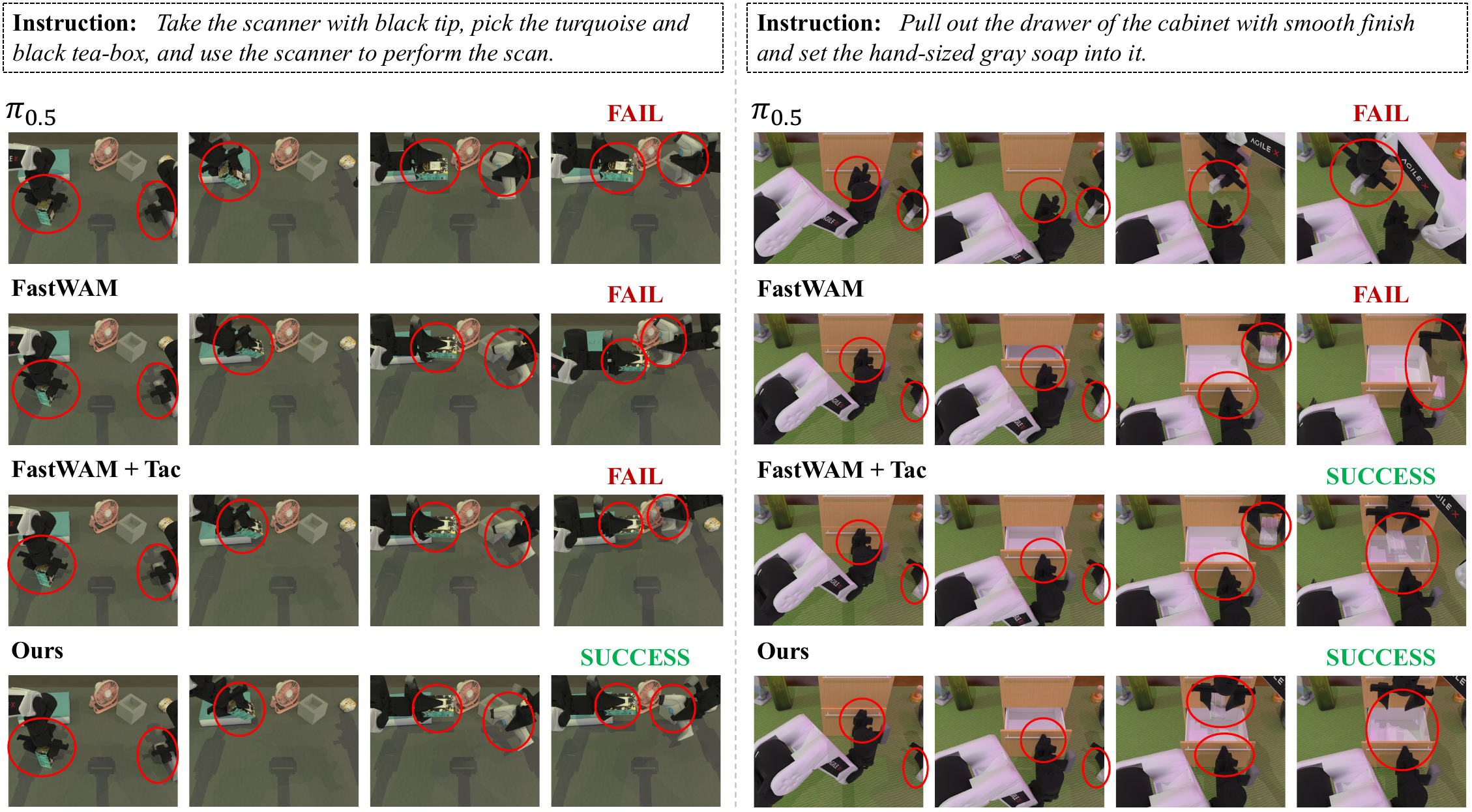}
    \caption{
\textbf{Qualitative comparisons on RoboTwin.}
Rollouts of $\pi_{0.5}$, Fast-WAM, Fast-WAM+\textsc{Tac}, and ME-Dex-1.0 on object scanning (left) and drawer placement (right). Frames progress from left to right.
}
    \label{fig:qualitative-robotwin}
\end{figure}

\vspace{-1mm}
ME-Dex-1.0  successfully completes both object scanning and drawer placement in the figure~\ref{fig:qualitative-robotwin} of shown rollouts, coordinating object grasping with subsequent manipulation.
In comparison, $\pi_{0.5}$ and Fast-WAM fail in both examples, while Fast-WAM+\textsc{Tac} completes drawer placement but fails the scanning task.
These examples highlight the advantage of ME-Dex-1.0 in completing manipulation sequences that require coordinated object interactions.


\begin{table}[t]
\centering
\caption{
\textbf{Tactile reconstruction across platforms.}
Metrics are averaged equally across subsets within each platform.
F1 scores are reported in percent and MAEs in source-specific normalized units.
}
\label{tab:tactile_reconstruction_summary}
\small
\setlength{\tabcolsep}{4pt}
\renewcommand{\arraystretch}{1.15}
\begin{tabularx}{\linewidth}{@{}l *{4}{>{\centering\arraybackslash}X}@{}}
\toprule
& \multicolumn{2}{c}{Uniformly sampled windows}
& \multicolumn{2}{c}{High-change windows} \\
\cmidrule(lr){2-3}
\cmidrule(l){4-5}
Platform
& \shortstack{Contact\\F1 $\uparrow$}
& \shortstack{Contact force\\MAE $\downarrow$}
& \shortstack{Transition\\F1 $\uparrow$}
& \shortstack{Transition delta\\MAE $\downarrow$} \\
\midrule
RoboTwin & 96.56 & 0.01797 & 89.73 & 0.01849 \\
DexJoCo  & 90.05 & 0.02807 & 68.73 & 0.04293 \\
ManiFeel & 99.69 & 0.05478 & 66.96 & 0.06899 \\
\bottomrule
\end{tabularx}
\end{table}

\subsection{Unified Tactile Representation}

We train a shared Unified Tactile Autoencoder on tactile data from RoboTwin, DexJoCo, and ManiFeel.
For each data source, we split trajectories into training and test sets at a ratio of 8 to 2 and combine the training sets for joint training.
Evaluation uses uniformly sampled windows and high-change windows.
For uniformly sampled windows, we report contact F1 and mean absolute error (MAE) of force values at contact nodes.
For high-change windows, we report contact-transition F1 and MAE of frame-to-frame force differences at contact onset and release locations.
Subset metrics are averaged equally within each platform, and force errors are computed in the normalized space of the corresponding source.

As shown in Table~\ref{tab:tactile_reconstruction_summary}, the shared model achieves contact F1 scores of 96.56\%, 90.05\%, and 99.69\% on RoboTwin, DexJoCo, and ManiFeel, respectively.
The contact and force reconstruction results support a shared representation across these sensing layouts.
Transition F1 scores of 89.73\%, 68.73\%, and 66.96\% further indicate that the representation retains information about contact onset and release, although the accuracy of these transitions varies across platforms.
The force MAEs should be interpreted within each source-specific normalized space.





\begin{figure}[!ht]
    \centering

    \includegraphics[width=\linewidth]{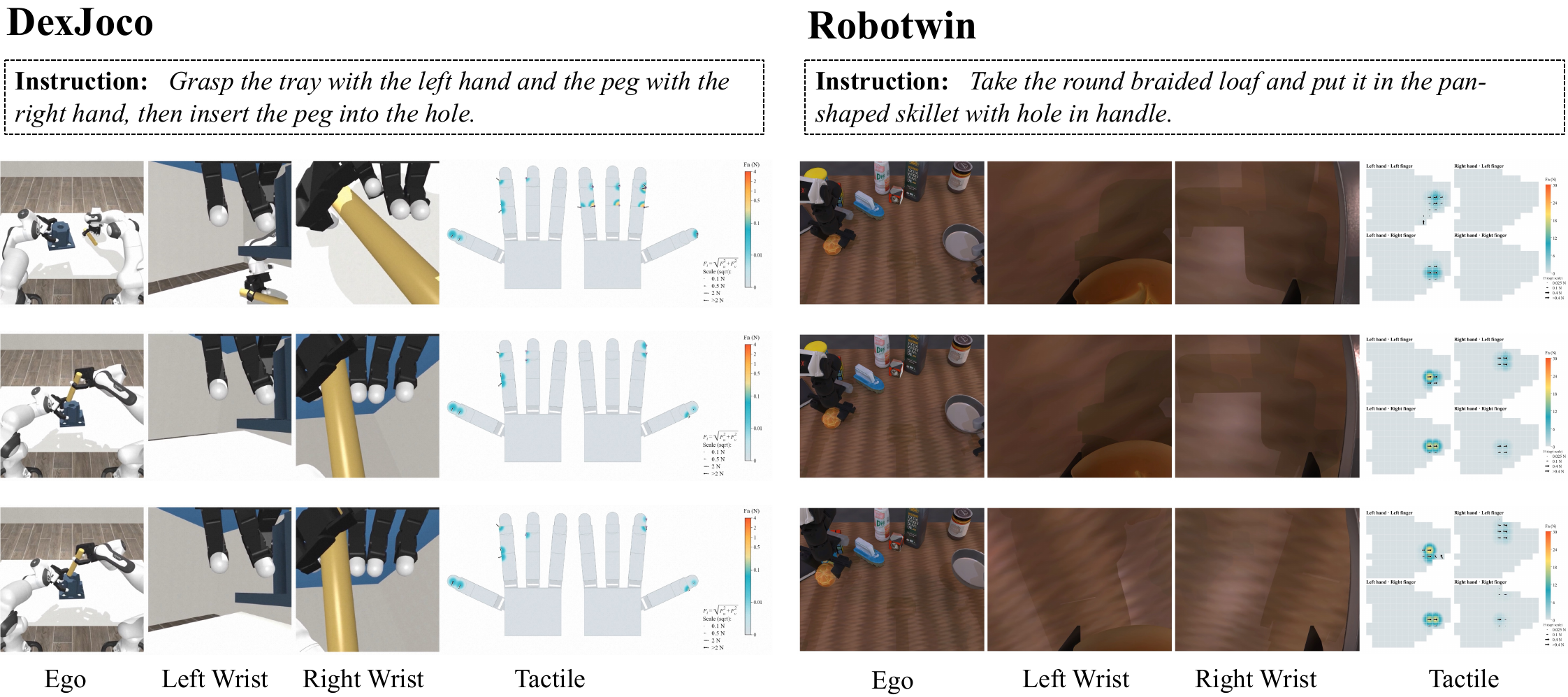}
    \caption{
        \textbf{ME-Dex-1.0 rollouts with tactile visualization in simulation.}
        Bimanual assembly on DexJoCo (left) and bread placement on
        RoboTwin (right). Each row shows ego and wrist-camera views
        alongside tactile sensor data at a selected timestep.
        Time progresses from top to bottom.
    }
    \label{fig:sim_tactile_rollouts}

    \vspace{6pt}

    \includegraphics[width=\linewidth]{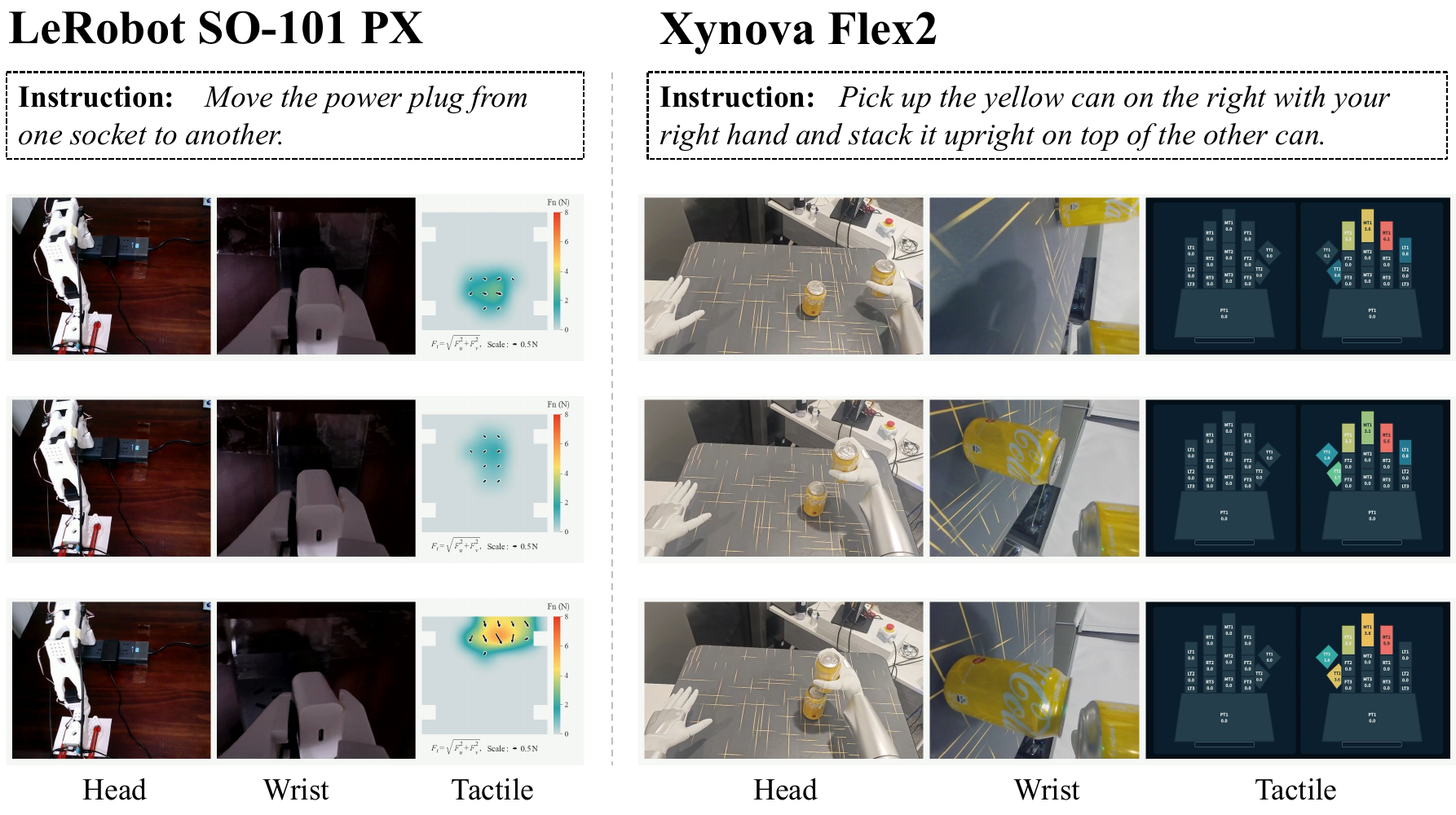}
    \caption{
        \textbf{ME-Dex-1.0 rollouts with tactile visualization on physical robots.}
        Power plug transfer with LeRobot SO-101 PX (left) and
        can stacking with Xynova Flex2 (right). Each row shows head
        and wrist-camera views alongside tactile sensor data
        at a selected timestep.
        Time progresses from top to bottom.
    }
    \label{fig:real_tactile_rollouts}

\end{figure}

\FloatBarrier

\subsection{Rollout Visualization with Tactile Data}

We visualize ME-Dex-1.0 rollouts through camera observations and tactile sensor data at selected timesteps. Figure~\ref{fig:sim_tactile_rollouts} presents examples from DexJoCo and RoboTwin, while Figure~\ref{fig:real_tactile_rollouts} shows execution on LeRobot SO-101 PX and Xynova Flex2. The tactile visualizations reveal changes in contact locations and force patterns during object interaction, complementing the corresponding camera views. These examples illustrate inference with multi-source heterogeneous tactile data across grippers and dexterous hands in both simulated and physical environments.





\section{Limitation and Future Work}
Pretraining of the unified tactile encoder remains limited by data scale and sensor diversity. We will expand cross-platform tactile data to cover more sensing layouts and contact patterns, and systematically evaluate generalization to unseen configurations. Current policies primarily use tactile feedback through observation updates and replanning between action chunks. We will explore tactile-driven local feedback control for high-frequency adjustments to jointly predicted actions during execution, aiming to improve responsiveness to contact changes and precision in fine manipulation.

\section{Conclusion}
We presented ME-X 1.0, a unified World Action Tactile Model that treats tactile states alongside video as future observations for action generation. Its Mixture-of-Transformers architecture brings together a Video Expert, a Tactile Expert, and an Action Expert for joint learning through flow matching. The experts exchange representations during joint prediction, connecting anticipated visual changes and contact dynamics with action generation. A unified tactile representation aligns multi-source heterogeneous tactile observations across embodiments and sensing layouts, while the Agentic Tactile Data Engine supplies paired training data through simulation. Experiments on RoboTwin, DexJoCo, and ManiFeel demonstrate improved manipulation performance with grippers and dexterous hands.

\section{Contributions}

\parab{Contributors.}
Xuancheng Zhang, Xuetao Liu, Qianying Tang, Jizhe Wang, Runsheng Wang, Zhijing Cheng, Bochen Lin, Haoran Wen, Ming Li, Kun Zhan, Yu Liu$^\dagger$

\vspace{6pt}
\noindent{
$^\dagger$Project Leader.}

\newpage

\bibliography{references}

@misc{luu2026manifeel,
  title         = {{{ManiFeel}: Benchmarking and Understanding Visuotactile Manipulation Policy Learning}},
  author        = {Luu, Quan Khanh and Zhou, Pokuang and Xu, Zhengtong and Zhang, Zhiyuan and Qiu, Qiang and She, Yu},
  year          = {2025},
  howpublished  = {arXiv preprint arXiv:2505.18472},
  eprint        = {2505.18472},
  archivePrefix = {arXiv},
  primaryClass  = {cs.RO}
}

@inproceedings{chen2025robotwin,
  title         = {{{RoboTwin 2.0}: A Scalable Data Generator and Benchmark with Strong Domain Randomization for Robust Bimanual Robotic Manipulation}},
  author        = {Chen, Tianxing and Chen, Zanxin and Chen, Baijun and Cai, Zijian and Liu, Yibin and Li, Zixuan and Liang, Qiwei and Lin, Xianliang and Ge, Yiheng and Gu, Zhenyu and Deng, Weiliang and Guo, Yubin and Nian, Tian and Xie, Xuanbing and Chen, Qiangyu and Su, Kailun and Xu, Tianling and Liu, Guodong and Hu, Mengkang and Gao, Huan-ang and Wang, Kaixuan and Liang, Zhixuan and Qin, Yusen and Yang, Xiaokang and Luo, Ping and Mu, Yao},
  booktitle     = {Proceedings of the 43rd International Conference on Machine Learning},
  year          = {2026}
}

@inproceedings{bi2025motus,
  title         = {{{Motus}: A Unified Latent Action World Model}},
  author        = {Bi, Hongzhe and Tan, Hengkai and Xie, Shenghao and Wang, Zeyuan and Huang, Shuhe and Liu, Haitian and Zhao, Ruowen and Feng, Yao and Xiang, Chendong and Rong, Yinze and Zhao, Hongyan and Liu, Hanyu and Su, Zhizhong and Ma, Lei and Su, Hang and Zhu, Jun},
  booktitle     = {Proceedings of the IEEE/CVF Conference on Computer Vision and Pattern Recognition (CVPR)},
  year          = {2026},
  pages         = {35101--35113}
}

@misc{higuera2026vtwm,
  title         = {{Visuo-Tactile World Models}},
  author        = {Higuera, Carolina and Arnaud, Sergio and Boots, Byron and Mukadam, Mustafa and Hogan, Francois Robert and Meier, Franziska},
  year          = {2026},
  howpublished  = {arXiv preprint arXiv:2602.06001},
  eprint        = {2602.06001},
  archivePrefix = {arXiv},
  primaryClass  = {cs.RO}
}

@misc{yuan2026fastwam,
  title         = {{{Fast-WAM}: Do World Action Models Need Test-time Future Imagination?}},
  author        = {Yuan, Tianyuan and Dong, Zibin and Liu, Yicheng and Zhao, Hang},
  year          = {2026},
  howpublished  = {arXiv preprint arXiv:2603.16666},
  eprint        = {2603.16666},
  archivePrefix = {arXiv},
  primaryClass  = {cs.CV}
}

@misc{motubrain2026,
  title         = {{{Motubrain}: An Advanced World Action Model for Robot Control}},
  author        = {{Motubrain Team} and Xiang, Chendong and Bao, Fan and Liu, Haitian and Tan, Hengkai and Bi, Hongzhe and Li, James and Liu, Jiabao and Pang, Jingrui and Jing, Kiro and Liu, Louis and Cai, Mengchen and Cui, Rongxu and Zhao, Ruowen and Wang, Runqing and Huang, Shuhe and Feng, Yao and Rong, Yinze and Wang, Zeyuan and Zhu, Jun},
  year          = {2026},
  howpublished  = {arXiv preprint arXiv:2604.27792},
  eprint        = {2604.27792},
  archivePrefix = {arXiv},
  primaryClass  = {cs.RO}
}

@misc{wang2026dexjoco,
  title         = {{{DexJoCo}: A Benchmark and Toolkit for Task-Oriented Dexterous Manipulation on {MuJoCo}}},
  author        = {Wang, Hanwen and Zhao, Weizhi and Wang, Xiangyu and Huang, Siyuan and Lin, He and Zheng, Boyuan and Xu, Rongtao and Wang, Gang and Mu, Yao and Wang, He and Fan, Lue and Li, Hongsheng and Zhang, Zhaoxiang and Tan, Tieniu},
  year          = {2026},
  howpublished  = {arXiv preprint arXiv:2605.16257},
  eprint        = {2605.16257},
  archivePrefix = {arXiv},
  primaryClass  = {cs.RO}
}

@misc{niu2026trex,
  title         = {{{T-Rex}: Tactile-Reactive Dexterous Manipulation}},
  author        = {Niu, Dantong and Liu, Zhuoyang and Wang, Zekai and Shao, Boning and Yin, Zhao-Heng and Pai, Anirudh and Sharma, Yuvan and Saravalle, Stefano and Zheng, Ruijie and Wang, Jing and Punamiya, Ryan and Xu, Mengda and Xie, Yuqi and Jiang, Yunfan and Fu, Letian and Kallidromitis, Konstantinos and Gioia, Matteo and Zhang, Junyi and Ge, Jiaxin and Feng, Haiwen and Galasso, Fabio and Zhan, Wei and Chan, David M. and Bai, Yutong and Herzig, Roei and Lei, Jiahui and Fei-Fei, Li and Goldberg, Ken and Malik, Jitendra and Abbeel, Pieter and Zhu, Yuke and Xu, Danfei and Fan, Linxi and Darrell, Trevor},
  year          = {2026},
  howpublished  = {arXiv preprint arXiv:2606.17055},
  eprint        = {2606.17055},
  archivePrefix = {arXiv},
  primaryClass  = {cs.RO}
}

@misc{bi2026motus2,
  title         = {{{Motus2}: A Self-Evolving General World Model for Dexterous Manipulation}},
  author        = {Bi, Hongzhe and Zhou, Zihao and Tang, Yihang and Pang, Jingrui and Huang, Shuhe and Liu, Haitian and Wang, Runqing and Huang, Shuai and Wang, Yichen and Cheng, Yiming and Zhao, Ruowen and Li, Zhenghua and Tan, Hengkai and Liu, Xiaolong and Wan, Jinhui and Liu, Jiabao and Zhao, Min and Bao, Fan and Zhu, Jun},
  year          = {2026},
  howpublished  = {arXiv preprint arXiv:2608.30237},
  eprint        = {2608.30237},
  archivePrefix = {arXiv},
  primaryClass  = {cs.RO}
}

@misc{wan2025wan,
  title         = {{{Wan}: Open and Advanced Large-Scale Video Generative Models}},
  author        = {{Team Wan} and Ang Wang and Baole Ai and Bin Wen and Chaojie Mao and Chen-Wei Xie and Di Chen and Feiwu Yu and Haiming Zhao and Jianxiao Yang and Jianyuan Zeng and Jiayu Wang and Jingfeng Zhang and Jingren Zhou and Jinkai Wang and Jixuan Chen and Kai Zhu and Kang Zhao and Keyu Yan and Lianghua Huang and Mengyang Feng and Ningyi Zhang and Pandeng Li and Pingyu Wu and Ruihang Chu and Ruili Feng and Shiwei Zhang and Siyang Sun and Tao Fang and Tianxing Wang and Tianyi Gui and Tingyu Weng and Tong Shen and Wei Lin and Wei Wang and Wei Wang and Wenmeng Zhou and Wente Wang and Wenting Shen and Wenyuan Yu and Xianzhong Shi and Xiaoming Huang and Xin Xu and Yan Kou and Yangyu Lv and Yifei Li and Yijing Liu and Yiming Wang and Yingya Zhang and Yitong Huang and Yong Li and You Wu and Yu Liu and Yulin Pan and Yun Zheng and Yuntao Hong and Yupeng Shi and Yutong Feng and Zeyinzi Jiang and Zhen Han and Zhi-Fan Wu and Ziyu Liu},
  year          = {2025},
  howpublished  = {arXiv preprint arXiv:2503.20314},
  eprint        = {2503.20314},
  archivePrefix = {arXiv},
  primaryClass  = {cs.CV}
}

@inproceedings{black2025pi05,
  title         = {{$\pi_{0.5}$: a Vision-Language-Action Model with Open-World Generalization}},
  author        = {Black, Kevin and Brown, Noah and Darpinian, James and Dhabalia, Karan and Driess, Danny and Esmail, Adnan and Equi, Michael Robert and Finn, Chelsea and Fusai, Niccolo and Galliker, Manuel Y. and Ghosh, Dibya and Groom, Lachy and Hausman, Karol and Ichter, Brian and Jakubczak, Szymon and Jones, Tim and Ke, Liyiming and LeBlanc, Devin and Levine, Sergey and Li-Bell, Adrian and Mothukuri, Mohith and Nair, Suraj and Pertsch, Karl and Ren, Allen Z. and Shi, Lucy Xiaoyang and Smith, Laura and Springenberg, Jost Tobias and Stachowicz, Kyle and Tanner, James and Vuong, Quan and Walke, Homer and Walling, Anna and Wang, Haohuan and Yu, Lili and Zhilinsky, Ury},
  booktitle     = {Proceedings of The 9th Conference on Robot Learning},
  year          = {2025},
  volume        = {305},
  pages         = {17--40},
  series        = {Proceedings of Machine Learning Research},
  publisher     = {PMLR}
}

@inproceedings{zheng2025xvla,
  title         = {{{X-VLA}: Soft-Prompted Transformer as Scalable Cross-Embodiment Vision-Language-Action Model}},
  author        = {Zheng, Jinliang and Li, Jianxiong and Wang, Zhihao and Liu, Dongxiu and Kang, Xirui and Feng, Yuchun and Zheng, Yinan and Zou, Jiayin and Chen, Yilun and Zeng, Jia and Zhang, Ya-Qin and Pang, Jiangmiao and Liu, Jingjing and Wang, Tai and Zhan, Xianyuan},
  booktitle     = {International Conference on Learning Representations},
  year          = {2026}
}

@article{chi2023diffusion,
  title         = {{Diffusion Policy: Visuomotor Policy Learning via Action Diffusion}},
  author        = {Chi, Cheng and Xu, Zhenjia and Feng, Siyuan and Cousineau, Eric and Du, Yilun and Burchfiel, Benjamin and Tedrake, Russ and Song, Shuran},
  journal       = {The International Journal of Robotics Research},
  year          = {2025},
  volume        = {44},
  number        = {10--11},
  pages         = {1684--1704},
}

@inproceedings{zhao2023act,
  title         = {{Learning Fine-Grained Bimanual Manipulation with Low-Cost Hardware}},
  author        = {Zhao, Tony Z. and Kumar, Vikash and Levine, Sergey and Finn, Chelsea},
  booktitle     = {Proceedings of Robotics: Science and Systems},
  year          = {2023},
}

@inproceedings{li2026lingbotva,
  title         = {{Causal World Modeling for Robot Control}},
  author        = {Li, Lin and Zhang, Qihang and Luo, Yiming and Yang, Shuai and Wang, Ruilin and Zhang, Luyao and Yu, Mingrui and Gao, Zelin and Xue, Nan and Zhou, Boyu and Zhu, Xing and Ding, Mingyu and Shen, Yujun and Xu, Yinghao},
  booktitle     = {Proceedings of Robotics: Science and Systems},
  year          = {2026},
}

@misc{zhou2026tau0wm,
  title         = {{{$\tau_0$-WM}: A Unified Video-Action World Model for Robotic Manipulation}},
  author        = {Zhou, Pengfei and Chen, Shengcong and Chen, Di and Wang, Jiaxu and Jin, Rongjun and Zhu, Bingwen and Pan, Yike and Gu, Songen and Wang, Kuanning and Nan, Shufeng and Qiu, Xingyu and Qiu, Chenhao and Yang, Pu and Cai, Yunuo and Gao, Jianxiong and Li, Yifan and Fu, Yanwei and Yue, Xiangyu and Chen, Zhi and Luo, Jianlan},
  year          = {2026},
  howpublished  = {arXiv preprint arXiv:2606.01027},
  eprint        = {2606.01027},
  archivePrefix = {arXiv},
  primaryClass  = {cs.RO}
}

@inproceedings{black2024pi0,
  title         = {{{$\pi_0$}: A Vision-Language-Action Flow Model for General Robot Control}},
  author        = {Black, Kevin and Brown, Noah and Driess, Danny and Esmail, Adnan and Equi, Michael Robert and Finn, Chelsea and Fusai, Niccolo and Groom, Lachy and Hausman, Karol and Ichter, Brian and Jakubczak, Szymon and Jones, Tim and Ke, Liyiming and Levine, Sergey and Li-Bell, Adrian and Mothukuri, Mohith and Nair, S and Vuong, Quan and Waluraj and Pertsch, Karl and Shi, Lucy Xiaoyang and Smith, Laura and Tanner, Jamesling, Anna and Wang, Haohuan and Zhilinsky, Ury},
  booktitle     = {Proceedings of Robotics: Science and Systems},
  year          = {2025},
}

@inproceedings{oneill2023openx,
  title         = {{Open X-Embodiment: Robotic Learning Datasets and {RT-X} Models}},
  author        = {{Open X-Embodiment Collaboration}},
  booktitle     = {2024 IEEE International Conference on Robotics and Automation (ICRA)},
  year          = {2024},
  pages         = {6892--6903},
}

@misc{huang2025tactilevla,
  title         = {{Tactile-{VLA}: Unlocking Vision-Language-Action Model's Physical Knowledge for Tactile Generalization}},
  author        = {Huang, Jialei and Wang, Shuo and Lin, Fanqi and Hu, Yihang and Wen, Chuan and Gao, Yang},
  year          = {2025},
  howpublished  = {arXiv preprint arXiv:2507.09160},
  eprint        = {2507.09160},
  archivePrefix = {arXiv},
  primaryClass  = {cs.RO}
}

@article{bi2025vlatouch,
  title         = {{{VLA-Touch}: Enhancing Vision-Language-Action Model With Dual-Level Tactile Feedback}},
  author        = {Bi, Jianxin and Ma, Kevin Yuchen and Hao, Ce and Shou, Mike Zheng and Soh, Harold},
  journal       = {IEEE Robotics and Automation Letters},
  year          = {2026},
  volume        = {11},
  number        = {7},
  pages         = {8487--8494},
}

@inproceedings{helmut2025farm,
  title         = {{Tactile-Conditioned Diffusion Policy for Force-Aware Robotic Manipulation}},
  author        = {Helmut, Erik and Funk, Niklas and Schneider, Tim and Farias, Cristiana de and Peters, Jan},
  booktitle     = {2026 IEEE International Conference on Robotics and Automation (ICRA)},
  year          = {2026}
}

@inproceedings{ayad2024imagine2touch,
  title         = {{{Imagine2Touch}: Predictive Tactile Sensing for Robotic Manipulation using Efficient Low-Dimensional Signals}},
  author        = {Ayad, Abdallah and R\"ofer, Adrian and Heppert, Nick and Valada, Abhinav},
  booktitle     = {ViTac 2024: Robot Embodiment through Visuo-Tactile Perception, ICRA Workshop},
  year          = {2024}
}

@misc{zheng2026omnivta,
  title         = {{{OmniVTA}: Visuo-Tactile World Modeling for Contact-Rich Robotic Manipulation}},
  author        = {Zheng, Yuhang and Gu, Songen and Zheng, Yupeng and Li, Weize and Zang, Yujie and Tian, Shuai and Li, Xiang and Hao, Ce and Gao, Chen and Liu, Si and Li, Haoran and Chen, Yilun and Yan, Shuicheng and Ding, Wenchao},
  year          = {2026},
  howpublished  = {arXiv preprint arXiv:2603.19201},
  eprint        = {2603.19201},
  archivePrefix = {arXiv},
  primaryClass  = {cs.RO}
}

@misc{zhou2026touchworld,
  title         = {{{TouchWorld}: A Predictive and Reactive Tactile Foundation Model for Dexterous Manipulation}},
  author        = {Zhou, Jianyi and Hong, Feiyang and Li, Yunhao and Zhao, Yicheng and Cen, Yongjue and Liu, Zirui and Huang, Jiakang and Chen, Zirui and Zhang, Ruiyang and Zhu, Weizhuo and Song, Xuhua and Yang, Shuo},
  year          = {2026},
  howpublished  = {arXiv preprint arXiv:2607.07287},
  eprint        = {2607.07287},
  archivePrefix = {arXiv},
  primaryClass  = {cs.RO}
}

@misc{huang2026vitacworld,
  title         = {{{ViTacWorld}: Scaling Visuo-Tactile World Models for Contact-Rich Robot Manipulation}},
  author        = {Huang, Yunao and Sang, Shiyu and Lu, Haotao and Ni, Suting and Wu, Shijie and Guo, Ziyang and Shi, Ye and Wang, Jingya},
  year          = {2026},
  howpublished  = {arXiv preprint arXiv:2607.22530},
  eprint        = {2607.22530},
  archivePrefix = {arXiv},
  primaryClass  = {cs.RO}
}

@inproceedings{yang2024unitouch,
  title         = {{Binding Touch to Everything: Learning Unified Multimodal Tactile Representations}},
  author        = {Yang, Fengyu and Feng, Chao and Chen, Ziyang and Park, Hyoungseob and Wang, Daniel and Dou, Yiming and Zeng, Ziyao and Chen, Xien and Gangopadhyay, Rit and Owens, Andrew and Wong, Alex},
  booktitle     = {Proceedings of the IEEE/CVF Conference on Computer Vision and Pattern Recognition (CVPR)},
  year          = {2024},
  pages         = {26340--26353}
}

@inproceedings{higuera2024sparsh,
  title         = {{Sparsh: Self-supervised touch representations for vision-based tactile sensing}},
  author        = {Higuera, Carolina and Sharma, Akash and Bodduluri, Chaithanya Krishna and Fan, Taosha and Lancaster, Patrick and Kalakrishnan, Mrinal and Kaess, Michael and Boots, Byron and Lambeta, Mike and Wu, Tingfan and Mukadam, Mustafa},
  booktitle     = {Proceedings of The 8th Conference on Robot Learning},
  year          = {2025},
  volume        = {270},
  pages         = {885--915},
  series        = {Proceedings of Machine Learning Research},
  publisher     = {PMLR}
}

@inproceedings{feng2025anytouch,
  title         = {{{AnyTouch}: Learning Unified Static-Dynamic Representation across Multiple Visuo-Tactile Sensors}},
  author        = {Feng, Ruoxuan and Hu, Jiangyu and Xia, Wenke and Gao, Tianci and Shen, Ao and Sun, Yuhao and Fang, Bin and Hu, Di},
  booktitle     = {International Conference on Learning Representations},
  year          = {2025}
}

@inproceedings{higuera2025sparshx,
  title         = {{Tactile Beyond Pixels: Multisensory Touch Representations for Robot Manipulation}},
  author        = {Higuera, Carolina and Sharma, Akash and Fan, Taosha and Bodduluri, Chaithanya Krishna and Boots, Byron and Kaess, Michael and Lambeta, Mike and Wu, Tingfan and Liu, Zixi and Hogan, Francois Robert and Mukadam, Mustafa},
  booktitle     = {Proceedings of The 9th Conference on Robot Learning},
  year          = {2025},
  volume        = {305},
  pages         = {105--123},
  series        = {Proceedings of Machine Learning Research},
  publisher     = {PMLR}
}

@article{wang2020tacto,
  title         = {{{TACTO}: A Fast, Flexible, and Open-Source Simulator for High-Resolution Vision-Based Tactile Sensors}},
  author        = {Wang, Shaoxiong and Lambeta, Mike and Chou, Po-Wei and Calandra, Roberto},
  journal       = {IEEE Robotics and Automation Letters},
  year          = {2022},
  volume        = {7},
  number        = {2},
  pages         = {3930--3937},
}

@article{si2021taxim,
  title         = {{{Taxim}: An Example-Based Simulation Model for GelSight Tactile Sensors}},
  author        = {Si, Zilin and Yuan, Wenzhen},
  journal       = {IEEE Robotics and Automation Letters},
  year          = {2022},
  volume        = {7},
  number        = {2},
  pages         = {2361--2368},
}

@inproceedings{si2024difftactile,
  title         = {{{DiffTactile}: A Physics-Based Differentiable Tactile Simulator for Contact-Rich Robotic Manipulation}},
  author        = {Si, Zilin and Zhang, Gu and Ben, Qingwei and Romero, Branden and Xian, Zhou and Liu, Chao and Gan, Chuang},
  booktitle     = {International Conference on Learning Representations},
  year          = {2024}
}

@inproceedings{brohan2023rt2,
  title         = {{RT-2: Vision-Language-Action Models Transfer Web Knowledge to Robotic Control}},
  author        = {Zitkovich, Brianna and Yu, Tianhe and Xu, Sichun and Xu, Peng and Xiao, Ted and Xia, Fei and Wu, Jialin and Wohlhart, Paul and Welker, Stefan and Wahid, Ayzaan and Vuong, Quan and Vanhoucke, Vincent and Tran, Huong and Soricut, Radu and Singh, Anikait and Singh, Jaspiar and Sermanet, Pierre and Sanketi, Pannag R. and Salazar, Grecia and Ryoo, Michael S. and Reymann, Krista and Rao, Kanishka and Pertsch, Karl and Mordatch, Igor and Michalewski, Henryk and Lu, Yao and Levine, Sergey and Lee, Lisa and Lee, Tsang-Wei Edward and Leal, Isabel and Kuang, Yuheng and Kalashnikov, Dmitry and Julian, Ryan and Joshi, Nikhil J. and Irpan, Alex and Ichter, Brian and Hsu, Jasmine and Herzog, Alexander and Hausman, Karol and Gopalakrishnan, Keerthana and Fu, Chuyuan and Florence, Pete and Finn, Chelsea and Dubey, Kumar Avinava and Driess, Danny and Ding, Tianli and Choromanski, Krzysztof Marcin and Chen, Xi and Chebotar, Yevgen and Carbajal, Justice and Brown, Noah and Brohan, Anthony and Arenas, Montserrat Gonzalez and Han, Kehang},
  booktitle     = {Proceedings of The 7th Conference on Robot Learning},
  year          = {2023},
  volume        = {229},
  pages         = {2165--2183},
  series        = {Proceedings of Machine Learning Research},
  publisher     = {PMLR}
}

@inproceedings{kim2024openvla,
  title         = {{OpenVLA: An Open-Source Vision-Language-Action Model}},
  author        = {Kim, Moo Jin and Pertsch, Karl and Karamcheti, Siddharth and Xiao, Ted and Balakrishna, Ashwin and Nair, Suraj and Rafailov, Rafael and Foster, Ethan P and Sanketi, Pannag R and Vuong, Quan and Kollar, Thomas and Burchfiel, Benjamin and Tedrake, Russ and Sadigh, Dorsa and Levine, Sergey and Liang, Percy and Finn, Chelsea},
  booktitle     = {Proceedings of The 8th Conference on Robot Learning},
  year          = {2025},
  volume        = {270},
  pages         = {2679--2713},
  series        = {Proceedings of Machine Learning Research},
  publisher     = {PMLR}
}

@misc{tian2026vtwam,
  title         = {{{VT-WAM}: Visual-Tactile World Action Model for Contact-Rich Manipulation}},
  author        = {Tian, Shuai and Zheng, Yupeng and Zheng, Yuhang and Gu, Songen and Zang, Yujie and Qin, Yuxing and Li, Weize and Li, Haoran and Ding, Wenchao and Zhao, Dongbin},
  year          = {2026},
  howpublished  = {arXiv preprint arXiv:2607.02503},
  eprint        = {2607.02503},
  archivePrefix = {arXiv},
  primaryClass  = {cs.RO}
}

@inproceedings{lin2024learning,
  title         = {{Learning Visuotactile Skills With Two Multifingered Hands}},
  author        = {Lin, Toru and Zhang, Yu and Li, Qiyang and Qi, Haozhi and Yi, Brent and Levine, Sergey and Malik, Jitendra},
  booktitle     = {2025 IEEE International Conference on Robotics and Automation (ICRA)},
  year          = {2025},
  pages         = {5637--5643},
}

@inproceedings{Li2026DECO,
  title         = {{{DECO}: Decoupled Multimodal Diffusion Transformer for Bimanual Dexterous Manipulation with a Plugin Tactile Adapter}},
  author        = {Xukun Li and Yu Sun and Lei Zhang and Bosheng Huang and Yibo Peng and Yuan Meng and Haojun Jiang and Shaoxuan Xie and Guocai Yao and Alois Knoll and Zhenshan Bing and Xinlong Wang and Zhenguo Sun},
  booktitle     = {Proceedings of the 43rd International Conference on Machine Learning},
  year          = {2026}
}

@inproceedings{wang2026hbridge,
  title         = {{{Hbridge}:H-shape bridging of heterogeneous experts for unified multimodal understanding and generation}},
  author        = {Wang, Xiang and Zhang, Zhifei and Zhang, He and Lin, Zhe and Zhou, Yuqian and Liu, Qing and Zhang, Shiwei and Li, Yijun and Liu, Shaoteng and Zheng, Haitian and Kuen, Jason and Wang, Yuehuan and Gao, Changxin and Sang, Nong.},
  booktitle     = {Proceedings of the IEEE/CVF Conference on Computer Vision and Pattern Recognition (CVPR)},
  year          = {2026},
  pages         = {14767-14778}
}

@article{Kim2025Openvlaoft,
title={{Fine-Tuning Vision-Language-Action Models:} Optimizing Speed and Success},
author={Kim, Moo Jin and Finn, Chelsea and Liang, Percy},
journal={arXiv preprint arXiv:2502.19645},
year={2025}
}

@inproceedings{wang2026vlaadapter,
  title   = {{VLA-Adapter}: An Effective Paradigm for
             Tiny-Scale Vision-Language-Action Model},
  author  = {Wang, Yihao and Ding, Pengxiang and Li, Lingxiao
             and Cui, Can and Ge, Zirui and Tong, Xinyang
             and Song, Wenxuan and Zhao, Han and Zhao, Wei
             and Hou, Pengxu and Huang, Siteng and Tang, Yifan
             and Wang, Wenhui and Zhang, Ru and Liu, Jianyi
             and Wang, Donglin},
  booktitle = {Proceedings of the AAAI Conference on Artificial Intelligence},
  volume  = {40},
  pages   = {18638--18646},
  year    = {2026},
}

@misc{ye2026dreamzero,
  title         = {World Action Models are Zero-shot Policies},
  author        = {Ye, Seonghyeon and Ge, Yunhao and Zheng, Kaiyuan
                   and Gao, Shenyuan and Yu, Sihyun and Kurian, George
                   and Indupuru, Suneel and Tan, You Liang
                   and Zhu, Chuning and Xiang, Jiannan and Malik, Ayaan
                   and Lee, Kyungmin and Liang, William and Ranawaka, Nadun
                   and Gu, Jiasheng and Xu, Yinzhen and Wang, Guanzhi
                   and Hu, Fengyuan and Narayan, Avnish and Bjorck, Johan
                   and Wang, Jing and Kim, Gwanghyun and Niu, Dantong
                   and Zheng, Ruijie and Xie, Yuqi and Wu, Jimmy
                   and Wang, Qi and Julian, Ryan and Xu, Danfei
                   and Du, Yilun and Chebotar, Yevgen and Reed, Scott
                   and Kautz, Jan and Zhu, Yuke and Fan, Linxi "Jim"
                   and Jang, Joel},
  year          = {2026},
  howpublished  = {arXiv preprint arXiv:2602.15922},
  eprint        = {2602.15922},
  archivePrefix = {arXiv},
  primaryClass  = {cs.RO}
}

@inproceedings{Wang20226wvlam,
  title         = {Unified vision-language-action model},
  author        = {Wang, Yuqi and Li, Xinghang and Wang, Wenxuan and Zhang, Junbo and Li, Yingyan Li and Chen, Yuntao and Wang, Xinlong and Zhang, Zhaoxiang},
  booktitle     = {International Conference on Learning Representations},
  year          = {2026}
}

@article{Zhun2025UWM,
title={{Unified world models:} Coupling video and action diffusion for pretraining on large robotic datasets},
author={Zhu, Chuning and Yu, Raymond and Feng, Siyuan and Burchfiel, Benjamin and Shah, Paarth and Gupta, Abhishek},
journal={arXiv preprint arXiv:2504.02792},
year={2025}
}

@article{Ye2026Gigaworld,
title={{GigaWorld-Policy:} An Efficient Action-Centered World--Action Model},
author={Ye, Angen and Wang, Boyuan Wang and Ni, Chaojun and Huang, Guan and Zhao, Guosheng and Li, Hao and Li, Hengtao and Li, Jie and Lv, Jindi and Liu, Jingyu and Cao, Min and Li, Peng and Deng, Qiuping and Mei, Wenjun and Wang, Xiaofeng and Chen, Xinze and Zhou, Xinyu and Wang, Yang and Chang, Yifan and Li, Yifan and Zhou, Yukun and Ye, Yun and Liu, Zhichao and Zhu Zheng},
journal={arXiv preprint arXiv:2603.17240},
year={2026}
}

@article{Ma2026DECOWAM,
title={{DECOWAM:} Decoupled Whole-Body World-Action Model for Legged Mobile Manipulation},
author={Ma, Siyuan and Zhang, Boshi and Zhang, Yutian and Wu, Qinglian and Zhai, Jiaqi and Wei, Dong and Yu, Qiaojun},
journal={arXiv preprint arXiv:2608.20114},
year={2026}
}

@inproceedings{zhao2025polytouch,
  title         = {{PolyTouch}: A Robust Multi-Modal Tactile Sensor for Contact-rich Manipulation Using Tactile-Diffusion Policies},
  author        = {Zhao, Jialiang and Kuppuswamy, Naveen
                   and Feng, Siyuan and Burchfiel, Benjamin
                   and Adelson, Edward},
  booktitle = {In 2025 IEEE International Conference on Robotics and Automation},
  pages   = {104-110},
  year    = {2025},
}

@inproceedings{zhu2025touchwild,
  title         = {{Touch in the Wild:} Learning Fine-Grained Manipulation with a Portable Visuo-Tactile Gripper},
  author        = {Zhu, Xinyue and Huang, Binghao and Li, Yunzhu},
  booktitle = {Advances in Neural Information Processing Systems},
  volume = {38},
  pages   = {153783-153812},
  year    = {2026},
}

@misc{heng2025vitacformer,
  title         = {{ViTacFormer}: Learning Cross-Modal Representation for Visuo-Tactile Dexterous Manipulation},
  author        = {Heng, Liang and Geng, Haoran and Zhang, Kaifeng
                   and Abbeel, Pieter and Malik, Jitendra},
  year          = {2025},
  howpublished  = {arXiv preprint arXiv:2506.15953},
  eprint        = {2506.15953},
  archivePrefix = {arXiv},
  primaryClass  = {cs.RO}
}

@article{raffel2020exploring,
  title={Exploring the limits of transfer learning with a unified text-to-text transformer},
  author={Raffel, Colin and Shazeer, Noam and Roberts, Adam and Lee, Katherine and Narang, Sharan and Matena, Michael and Zhou, Yanqi and Li, Wei and Liu, Peter J},
  journal={Journal of machine learning research},
  volume={21},
  number={140},
  pages={1--67},
  year={2020}
}

\appendix


\end{document}